\documentclass[12pt]{article}
\usepackage[T1]{fontenc}
\usepackage{lmodern}
\usepackage[left=1in,right=1in,top=1in,bottom=1in]{geometry}
\usepackage{amsmath,amsfonts,amssymb,amsthm}
\usepackage{graphicx,booktabs,float,multirow,longtable,array,tabularx}
\usepackage{subcaption,titlesec,authblk}
\usepackage[dvipsnames]{xcolor}
\definecolor{revisiongreen}{RGB}{0,120,0}

\usepackage{colortbl}
\usepackage{tikz}
\usetikzlibrary{arrows.meta,positioning,shapes.geometric,calc}
\usepackage[most]{tcolorbox}
\usepackage{xurl,needspace,multicol}
\usepackage[section]{placeins}
\usepackage[normalem]{ulem}
\usepackage{makecell,enumitem,adjustbox}
\usepackage{natbib}
\usepackage{hyperref}
\hypersetup{colorlinks=true,allcolors=black}
\setcitestyle{numbers,square}
\newcommand{\PTR}{PTR}
\newcommand{\OPTN}{OPTN}

\newcommand{\KDPI}{KDPI}

\newcommand{\RF}{Random Forest}
\newcommand{\XGB}{XGBoost}
\definecolor{arbGreen}{HTML}{387627}
\definecolor{arbGold}{HTML}{B68A00}
\definecolor{arbBlue}{HTML}{347FB8}
\definecolor{arbRed}{HTML}{BD513F}
\definecolor{dblue}{rgb}{0,0,0}

\title{When Should AI Decide, Defer, and Deviate? Responsible Automation for Deceased Donor Kidney Allocation}
\author[1]{Madeleine Pollack}
\author[1]{Thomas Daillak}
\author[2,3]{Mehmet Toner}
\author[2,3]{Korkut Uygun}
\author[2,3]{Heidi Yeh}
\author[1]{Swati Gupta}

\affil[1]{\small Massachusetts Institute of Technology, Cambridge, MA 02139}
\affil[2]{\small Massachusetts General Hospital, Boston, MA 02114}
\affil[3]{\small Harvard Medical School, Boston, MA 02115}

\date{}

\begin{document}
\maketitle

\begin{abstract}
We ask when an AI trained on historical decisions should be allowed to recommend \textit{decisions}, \textit{defer} to human experts, or in other cases, flag concerning patterns in data and \textit{deviate} from past practice. This can be challenging because past decisions record what \textit{was} done rather than what \textit{should have been} done, and where data is sparse, a model's predictions may be too uncertain to act on.
In this work, we provide a framework for addressing this problem using a case study in deceased-donor kidney allocation.
Nearly 30\% of kidneys recovered from deceased donors in the United States are never transplanted, largely because they are hard to place through the standard waitlist sequence. One can use AI to flag kidneys at risk of delayed or failed placement for out-of-sequence (OOS) allocation, but this comes at the cost of bypassing candidates higher on the waitlist. {Thus, OOS allocation should only be used when the organ is demonstrably at high risk of discard on the standard sequence.}
Using Organ Procurement and Transplantation Network match-run data from 2016--2019 and 2023--2025 (excluding COVID-19-disrupted years), we build the \textit{Arbitrator} framework, which combines various interpretable models and acts only when the uncertainty in their predictions cannot change the preferred action. Our Arbitrator resolves 93.5\% of organ placement decisions using explicit medical criteria learned from data, and defers the remaining 6.5\% to a human or a more complex model. We show that the Arbitrator correctly identifies 70.1\% of historically discarded low-KDPI kidneys as OOS candidates, projecting that up to 43.9\% of historically discarded kidneys could be flagged for expedited placement and potentially transplanted.  We then exploit the interpretability of the Arbitrator to create an agentic audit comparing the learned rules to published clinical literature in order to flag decision-making patterns that are contentious. We discuss various future directions that stem from this advancement due to AI.
\end{abstract}


\section{Introduction}
In 2023, nearly 30\% of kidneys recovered from deceased donors in the United States were not transplanted despite a persistent and overwhelming shortage of organs for patients with kidney failure \citep{lentine2025}.
The problem is not simply that all non-transplanted kidneys are medically unsuitable; rather, many of these organs are likely transplantable upon recovery \citep{aubert2019}, but deteriorate over the length of time required for an organ procurement organization (OPO) to find a suitable match for the organ \citep{bradbrook2025,wood2025,barah2022}.

Under \OPTN{} policy, a deceased donor kidney in the United States is offered sequentially to compatible candidates in a priority sequence, or \emph{match run}, determined by national allocation rules \citep{hrsa_continuous_distribution_kidney_pancreas_2025}. This sequence is designed to make the organ offer process systematic, fair, and auditable.  However, if OPOs believe that sequential offering is unlikely to succeed by the time the organ expires, they may choose to expedite placement via out-of-sequence (OOS) placement \citep{king2022,hrsa_allocation_out_of_sequence}. This creates a tradeoff: adhering to the match run risks nonuse of hard-to-place organs, while OOS placement bypasses candidates with higher priority, undermining the fairness the match run is designed to ensure. The concern that OOS allocation processes may be unfair is not unmerited: recent work has documented socioeconomic differences among recipients of OOS kidneys \citep{king2022,king2023,mohan2025}, and more work is needed to ensure OOS allocation is not exacerbating inequities in the healthcare system.
A useful decision aid must therefore distinguish organs for which OOS allocation prevents nonuse from those easily matchable organs for which the standard sequence should be preserved.

Machine learning (ML) and, more broadly, artificial intelligence (AI) are attractive tools for this problem because donor profiles are high dimensional and organ acceptance reflects nonlinear interactions between clinical characteristics. Predictive models have already been developed for kidney discard, delayed placement, and offer acceptance \citep{massie2010,zhou2019,barah2021,bertsimas2017,wey2018}. But the barrier to responsible deployment is not predictive performance alone. In a qualitative study of 23 healthcare professionals considering AI-driven clinical decision support for resource allocation, Y{\"u}ksel Elgin and Elgin found recurring concerns about the cost of increased efficiency in terms of equity, transparency and explainability.
Participants also stressed the need for additional review and manual oversight of AI decisions and emphasized that AI should be used for decision support rather than a replacement for clinical expertise \citep{yukselelgin2024}.

\begin{figure*}[t]
\centering
\resizebox{0.95\textwidth}{!}{\begin{tikzpicture}[
  font=\footnotesize,
  stage/.style={
    draw,
    rounded corners=2pt,
    align=center,
    minimum height=1.0cm,
    text width=1.6cm,
    inner ysep=3pt,
    fill=gray!4
  },
  model/.style={stage, text width=1.8cm, fill=blue!4},
  certnode/.style={stage, text width=2.1cm},   arb/.style={
    draw,
    rounded corners=2pt,
    align=left,
    minimum height=1.9cm,
    text width=3.3cm,
    inner ysep=4pt,
    fill=orange!5
  },
  tier/.style={stage, text width=1.6cm, minimum height=0.9cm},
  auditnode/.style={stage, text width=2.4cm, minimum height=1.1cm, fill=purple!4},
  outnode/.style={
    draw,
    rounded corners=2pt,
    align=center,
    minimum height=0.9cm,
    text width=2.4cm,
    inner ysep=3pt
  },
  arr/.style={-{Latex[length=1.9mm]}, semithick},
  lbl/.style={font=\footnotesize\itshape, inner sep=1pt}
]

\node[stage, text width=1.3cm] (x) at (0,0) {Donor\\features};

\node[model] (cart) at (2.6,1.0)  {\textbf{CART}};
\node[model] (rf)   at (2.6,-1.0) {\textbf{FIRE-}\\\textbf{sparsified RF}};

\node[certnode] (certc) at (5.5,1.0)  {certify EM\\certify \textit{not} EM\\or \textbf{abstain}};
\node[certnode] (certr) at (5.5,-1.0) {certify EM\\certify \textit{not} EM\\or \textbf{abstain}};

\node[lbl, align=center, anchor=south, yshift=8pt]
  at (cart.north |- certc.north)
  {Interpretable models \&\\organ-specific uncertainty intervals};

\coordinate (merge) at (7.05,0);

\node[arb] (arb) at (9.75,0) {
  \textbf{Arbitrator}\\[2pt]
  certified by either and no conflict
  \quad$\rightarrow$ \textbf{resolved}\\
  otherwise $\rightarrow$ \textbf{unresolved}
};

\node[tier] (res)   at (13.3,1.0)  {Resolved\\\textbf{93.5\%}};
\node[tier] (unres) at (13.3,-1.6) {Unresolved\\\textbf{6.5\%}};

\node[auditnode] (audit) at (16.15,1.0)
  {\textbf{Auditor}\\{\scriptsize clinical references\\+ rule interval}};

\node[outnode, fill=green!6]   (decide)  at (19.9,2.3)
  {\textbf{Decide}\\{\scriptsize aligned, confident}};
\node[outnode, fill=orange!10] (deviate) at (19.9,1.0)
  {\textbf{Deviate}\\{\scriptsize biased or obsolete}};
\node[outnode, fill=yellow!8]  (defer)   at (19.9,-1.6)
  {\textbf{Defer}\\organ-level review};

\draw[arr] (x.east) -- ++(0.35,0) |- (cart.west);
\draw[arr] (x.east) -- ++(0.35,0) |- (rf.west);
\draw[arr] (cart) -- (certc);
\draw[arr] (rf) -- (certr);

\draw[semithick] (certc.east) -- (merge |- certc.east) -- (merge);
\draw[semithick] (certr.east) -- (merge |- certr.east) -- (merge);
\draw[arr] (merge) -- (arb.west);

\draw[arr] (arb.east) -- ++(0.4,0) |- (res.west);
\draw[arr] (arb.east) -- ++(0.4,0) |- (unres.west);

\draw[arr] (res) -- (audit);
\coordinate (bus) at ([xshift=0.5cm]audit.east);
\draw[arr] (audit.east) -- (bus) |- (decide.west);
\draw[arr] (audit.east) -- (deviate.west);
\draw[arr] (bus) |- ([yshift=5pt]defer.west)
  node[lbl, pos=0.3, left=3pt, align=right] {weak evidence or\\needs context};

\draw[arr] ([yshift=-5pt]unres.east) -- ([yshift=-5pt]defer.west);

\coordinate (tip)  at ([yshift=-3pt]certr.south);
\coordinate (ctop) at (0,-2.25);   \coordinate (cbot) at (0,-3.6);    \draw[gray!60, fill=gray!3, rounded corners=3pt, line join=round]
  (x.west |- ctop) -- ([xshift=-0.25cm]tip |- ctop)
  -- (tip)
  -- ([xshift=0.25cm]tip |- ctop) -- (defer.east |- ctop)
  -- (defer.east |- cbot) -- (x.west |- cbot) -- cycle;

\begin{scope}[shift={(1.6,-2.92)}, font=\footnotesize]

  \def\dx{0}
  \draw[gray!45] (\dx-1.75,0) -- (\dx+1.75,0);
  \draw[dashed,black,semithick] (\dx,-0.22) -- (\dx,0.22);
  \node[black,below=-1pt] at (\dx,-0.22) {$\tau_\beta$};
  \draw[arbGreen,line width=1.5pt] (\dx+0.35,0) -- (\dx+1.25,0);
  \draw[arbGreen,line width=1.5pt] (\dx+0.35,-0.10) -- (\dx+0.35,0.10);
  \draw[arbGreen,line width=1.5pt] (\dx+1.25,-0.10) -- (\dx+1.25,0.10);
  \node[arbGreen,above=-1pt] at (\dx+0.80,0.22) {certify on-sequence};

  \def\dx{5.5}
  \draw[gray!45] (\dx-1.75,0) -- (\dx+1.75,0);
  \draw[dashed,black,semithick] (\dx,-0.22) -- (\dx,0.22);
  \node[black,below=-1pt] at (\dx,-0.22) {$\tau_\beta$};
  \draw[arbGreen,line width=1.5pt] (\dx-1.25,0) -- (\dx-0.35,0);
  \draw[arbGreen,line width=1.5pt] (\dx-1.25,-0.10) -- (\dx-1.25,0.10);
  \draw[arbGreen,line width=1.5pt] (\dx-0.35,-0.10) -- (\dx-0.35,0.10);
  \node[arbGreen,above=-1pt] at (\dx-0.80,0.22) {certify out-of-sequence};

  \def\dx{11.0}
  \draw[gray!45] (\dx-1.75,0) -- (\dx+1.75,0);
  \draw[dashed,black,semithick] (\dx,-0.22) -- (\dx,0.22);
  \node[black,below=-1pt] at (\dx,-0.22) {$\tau_\beta$};
  \draw[arbRed,line width=1.5pt] (\dx-0.55,0) -- (\dx+0.55,0);
  \draw[arbRed,line width=1.5pt] (\dx-0.55,-0.10) -- (\dx-0.55,0.10);
  \draw[arbRed,line width=1.5pt] (\dx+0.55,-0.10) -- (\dx+0.55,0.10);
  \node[arbRed,above=-1pt] at (\dx,0.22) {abstain};

  \node[anchor=west, align=left, font=\scriptsize\itshape, gray!45!black]
    at (13.4,0.02) {
    A wide interval can still certify;\\
    a narrow one straddling $\tau_\beta$ cannot.
  };

\end{scope}
\end{tikzpicture}}

\caption{\textbf{Our uncertainty-aware, auditable Arbitrator model.}
Each interpretable model certifies an action only when its confidence interval lies entirely on one side of $\tau_\beta$. Organs with certified, non-conflicting actions are resolved; all others are deferred. An auditor (expert or language model) checks each resolved rule against clinical references and recommends deciding, deviating, or further deferring. \XGB{} handles deferred organs.}
\label{fig:framework}
\end{figure*}
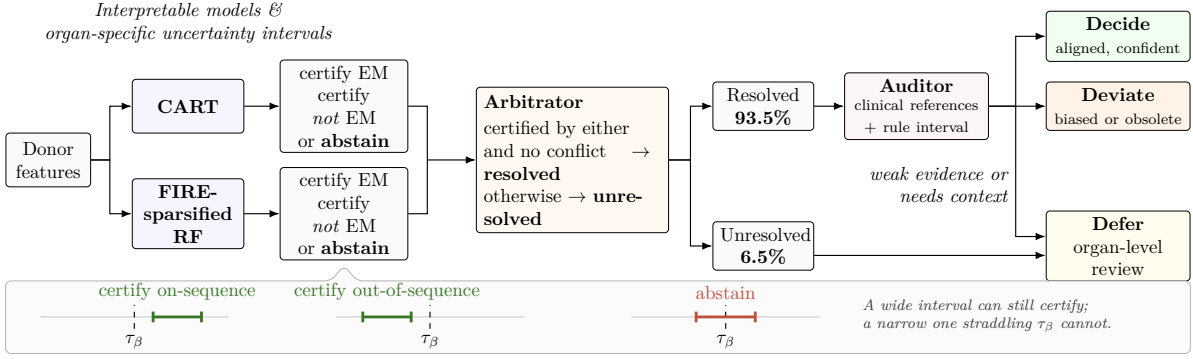

It is evident that in high-stakes decisions, a model with high predictive accuracy on past data does not imply that the model's learned behaviors are always worth replicating. AI models trained on historical outcomes can learn and reproduce the bias of the dataset when used for future high-stakes decisions \citep{obermeyer2019}, and historical organ allocation decisions themselves show socioeconomic disparities \citep{king2022,mohan2025}. Beyond bias, decision guidelines can also become outdated with new research and evolving transplantation policy, meaning that a good decision using the best available information ten years ago may differ from a good decision using the best available information today.
These concerns motivate a shift from prioritizing accuracy in prediction to generating uncertainty-aware prescription. That is, AI models must support decisions about not just \textit{how} to intervene, but also \textit{whether} to intervene based on the model's limitations.
Our work seeks to incorporate three relevant components when using AI to recommend kidneys for out-of-sequence allocation: interpretability, uncertainty-awareness, and validation from external evidence. These components are integrated sequentially in our framework in Figure \ref{fig:framework}.

First, we use interpretable\footnote{We define an \textit{interpretable model} as one whose predictions are defined by a \textit{decision rule}, e.g., ``if KDPI $\leq x$ and creatinine $\leq y$ and patient age is between $z_1$ and $z_2$, then predict class $C$ with probability $p$.''} ML models to predict the probability that a deceased-donor kidney will be accepted by a patient on the match run within 100 offers. We call this outcome early-matchability (EM), and we recommend that OOS allocation be considered when an organ is not likely to be EM. Given a relative preference $\beta$ for maintaining the match run versus routing high-risk organs OOS, the optimal action is to send an organ OOS if its EM probability is at most $\tau_\beta$.

However, a given ML/AI prediction is not guaranteed to be accurate, and different predictions can come with different levels of certainty based on the organ's features. Our models are \textit{uncertainty-aware}, which means that the model abstains from making a decision if its confidence interval for the EM probability contains $\tau_\beta$; that is, if the uncertainty in the estimated EM probability could change the recommended action. Otherwise, the model \textit{certifies} either the on-sequence (OS) action (the full confidence interval lies above $\tau_\beta$) or the OOS action (the full confidence interval lies below $\tau_\beta$).
Because we have multiple interpretable ML models, we use the \textit{Arbitrator} decision rule to resolve their recommendations, submitting either a certified action or deferring the organs for additional review, which, in practice, can be done by humans. In this work, we defer the decision to a more powerful, but ``black-box'' XGBoost model.

Finally, because each model outputs a set of interpretable decision rules, we audit each resolved recommendation for alignment with published medical literature instead of blindly replicating historical decisions. This helps us identify problematic trends learned by AI models from historic data. Our agentic auditor can make one of three decisions: (1) provisionally retain the proposed optimal action (decide), (2) propose considering the \textit{opposite} action due to medical research (deviate), or (3) request additional review for the model rule and recommendation due to insufficient evidence (defer). This model intends to maintain the parts of historical decision-making worth replicating, while allowing humans to intervene where historical biases, new research, or noise may render a specific recommendation too risky to automate.

\section{Materials and Methods}
We describe how we construct the study cohort, fit and calibrate prediction models, quantify uncertainty, combine certified decisions through the Arbitrator, and audit the resulting rules against clinical literature.

\subsection{Data sources and cohort construction}

We use \OPTN{} \PTR{} match-run data linked to the Standard Transplant Analysis and Research (STAR) deceased donor and waitlist registries for the pooled years 2016--2019 and 2023--2025 to predict whether a deceased donor kidney will be accepted at original PTR sequence number 100 or earlier\footnote{This choice was made at the start of this study to remove COVID-19 effects. The 2025 data extend through June 2025.}. The \PTR{} dataset records offers made to compatible candidates and their disposition, including bypass, refusal, provisional acceptance, and acceptance. We remove bypassed and date-specific operational-refusal offer rows and exclude donors whose match runs contain out-of-sequence or expedited-placement codes so that the modeled cohort reflects standard on-sequence allocation to the extent observable in the data. We also exclude dual-kidney allocation and selected multiorgan listing pathways whose allocation processes differ from kidney-alone placement. Detailed refusal codes, missing-value handling, feature construction, and leakage exclusions are reported in the Appendix, \S~\ref{sec:si-cohort}--\ref{sec:si-models}.

Kidneys from the same donor are kept within the same data split. Donors are assigned 70\% to training, 15\% to validation, and 15\% to test, stratified by \KDPI{} decile and the number of early-accepted kidneys per donor. The final analysis contains 100,674 training organs, 21,573 validation organs, and 21,574 test organs; the test set contains 14,127 organs (65.5\%) accepted within 100 offers and 7,447 (34.5\%) not accepted within that cutoff. Training data determine model hyperparameters and probability calibration, validation data determine the uncertainty intervals, and test outcomes are used only for final evaluation.

\subsection{Prediction models and calibration}

We evaluate six predictors: a thresholded \KDPI{} rule for organs with \KDPI{} $\le85\%$, a calibrated \KDPI{} rule, logistic regression, CART, \RF{}, and \XGB{}. Logistic regression uses five clinically relevant predictors: creatinine, age, diabetes history, \KDPI{}, and pediatric status. The CART, \RF{}, and \XGB{} models are trained on the full donor feature set including demographics, clinical history, serologies, kidney-specific measurements, biopsy variables, and others. For logistic regression, CART, \RF{}, and \XGB{}, isotonic calibration is fitted to donor-grouped out-of-fold training predictions. Calibrated \KDPI{} uses isotonic regression fitted to training outcomes using $1-\mathrm{KDPI}/100$ as the predictor, with KDPI expressed as a percentage. The raw \KDPI{} baseline remains uncalibrated.

\subsection{Local uncertainty}
Each interpretable model outputs for a specific organ a probability that the organ is EM, a 99\% confidence interval for that point prediction, and an interpretable rule that explains why that organ is given that point prediction and confidence level. Here we explain how these confidence intervals are generated for each model. More details can be found in the Appendix, \S~\ref{sec:fire-budget}.
For CART, each rule is given by its terminal leaves, so each kidney matches exactly one leaf. We attach the standard Wilson score interval at nominal confidence level $0.99$ \citep{wilson1927probable} for the fraction of validation organs in each leaf of the trained CART model that are early matchable.

In contrast, Random Forest is not naturally an interpretable model, so we turn to Fast Interpretable Rule Extraction (FIRE) \citep{liu2023fire} to select a compact set of rules from our trained \RF{}, which we call our SparseRF rule set.
We choose to keep at most 128 rules to allow sufficient coverage, which yields 80 selected rules.
Each selected rule receives a nominal 99\% Wilson interval from matching validation organs. For an organ satisfying multiple validation-supported rules, we take the convex hull of the Wilson intervals for each applicable rule (i.e., we take the lowest lower bound and the highest upper bound out of all of the applicable rules). Consequently, SparseRF certifies an action only when \textit{all} applicable supported rules certify that same action; otherwise it abstains. If no rule covers an organ, we apply confidence interval $[0,1]$.

\subsection{Cost-aware certification and arbitration}
We assume that the costs of unfairly routing an organ OOS versus possibly discarding a usable organ are nonuniform, so we assign a tunable cost of $\beta$ to the former and a cost of 1 to the latter. If an organ's true EM probability is $p$, then the expected costs associated with keeping an organ on-sequence versus routing it out of sequence are, respectively,
\begin{equation}
C_{\text{OS}}(p)=1-p, \qquad C_{\text{OOS}}(p)=\beta p.
\end{equation}
The optimal action then keeps the organ on-sequence when $p\ge \tau_\beta=1/(1+\beta)$, and routes the organ out-of-sequence otherwise. The primary operating point for this paper is $\beta=0.5$.

In our setting, we do not know the true value $p$. Instead, we have its upper and lower bounds with 99\% confidence on the validation set given by the Wilson score interval $[L, U]$. We say a model certifies an on-sequence (OS) decision if $L>\tau_\beta$, certifies out-of-sequence (OOS) if $U<\tau_\beta$, and abstains otherwise. Equivalently, a model certifies an action only when that action remains optimal across the entire uncertainty interval.

The Arbitrator is tasked with resolving the certifications of each of the interpretable models---in our case CART and SparseRF. If both models certify the same action, the Arbitrator uses that action. If exactly one model certifies and the other abstains, the certified action is used. If both abstain or certify conflicting actions, the Arbitrator defers the kidney for organ-level review, represented here by the preference-weighted \XGB{} plug-in decision, which uses $\tau_\beta$ as the decision threshold.

\subsection{LLM-assisted clinical rule audit}
We conducted an exploratory literature-assisted audit of the 113 certified contributing rules (35 condensed CART rules\footnote{We collapse subtrees whose leaves certify the same action for the sake of conciseness in our analysis.} out of 63 originally certified leaves and 78 SparseRF rules) at $\beta=0.5$. We asked a GPT-6-based OpenAI Codex agent to identify 10--20 papers to inform decide/defer/deviate assessments of certified rules, explicitly prioritizing creatinine-driven OOS recommendations for AKI-related reconsideration. The agent selected 18 research papers and one OPTN monitoring report and one clinical guidance document for comparison to the learned rules. The author instructions, agent-selected literature, agent-developed assignment criteria, and limitations are described in the Appendix, \S~\ref{sec:si-audit-methods}.

\section{Results}

We first compare CART, Random Forest and gradient-boosted trees (\XGB{}) with logistic regression and raw and calibrated \KDPI{}\footnote{The Kidney Donor Profile Index (KDPI) is a score from 0\% to 100\% estimating the percentage of recovered kidneys with lower transplant risk than this kidney. Lower scores proxy higher-functioning kidneys.} benchmarks, using donor-only predictors.
We evaluate all models on a held-out test set of 21,574 kidneys from the pooled 2016--2019 and 2023--2025 cohort. Of these, 65.5\% were early-matchable (EM). Because our goal is to identify transplantable kidneys at risk of discard rather than just identifying poor quality kidneys for OOS allocation, we will often report results on a ``low-KDPI'' subset of test organs, which we define as kidneys with \KDPI{} below $ 85\%$\footnote{This 85\% threshold is meaningful because \OPTN{} policy requires candidates to provide prior written informed consent to receive offers of kidneys with KDPI above 85\%~\citep{optn_policy_5_3_c}.}. This organ subset serves as a proxy for the pool of organs that could potentially be saved if OOS allocation is used.

We first show the benefit of ML over clinical heuristics for identifying non-EM organs. We then show that two interpretable models resolve most routing decisions with explicit rules, and we quantify the trade-off between interpretability, average decision cost, and burden on human decision-makers. These high-performing, interpretable base models are inputs into the Arbitrator decision rule, which resolves whether a particular organ has sufficient evidence to be a candidate for automation either OS or OOS. Finally, we illustrate how the resulting rules can be compared to clinical evidence using an agentic AI auditor to ensure that recommended actions align with clinical practice.

\subsection{Comparative Performance of ML Models}

\begin{figure}[ht]
\centering
\includegraphics[width=.9\linewidth]{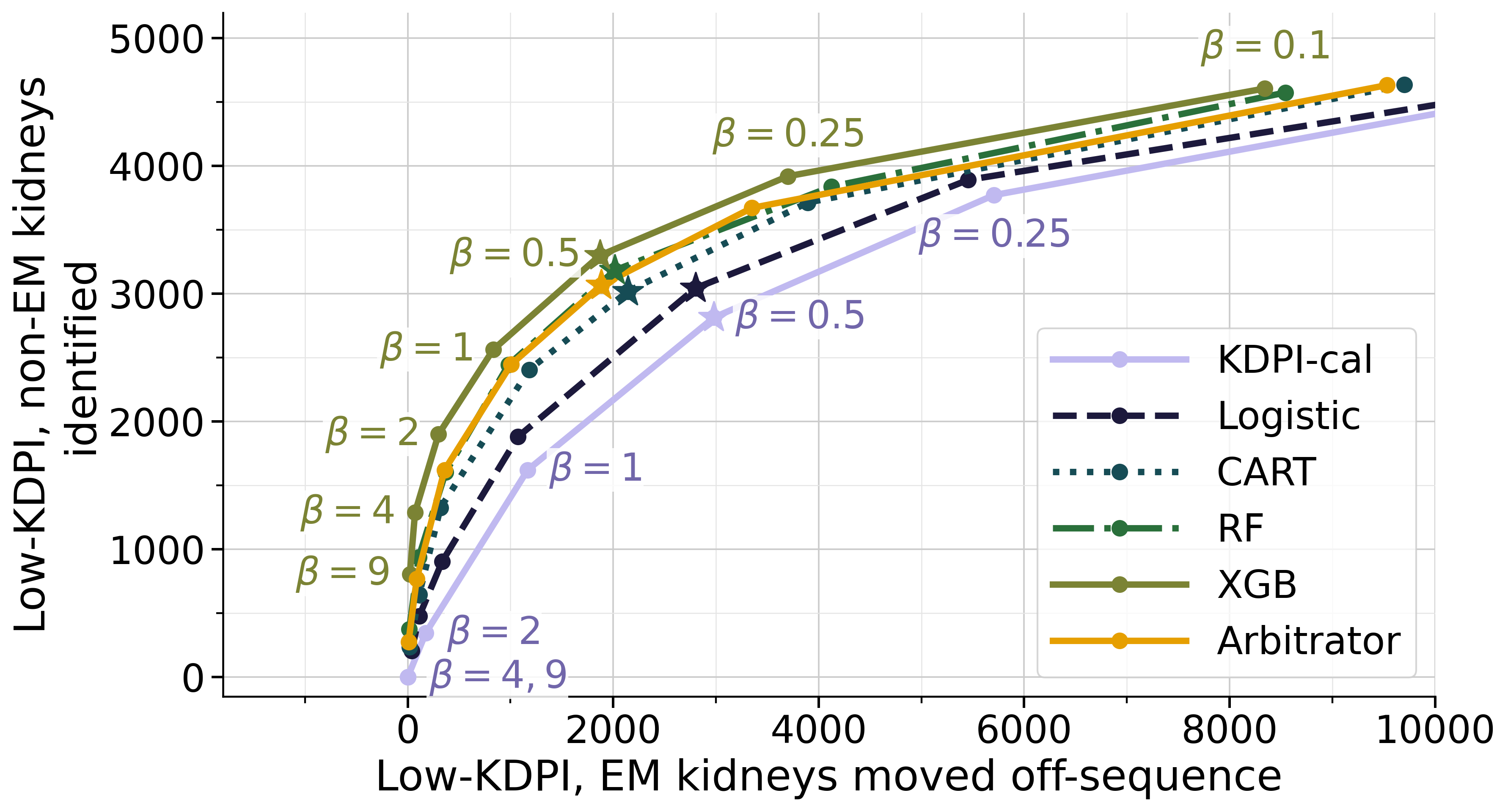}
\caption{\textbf{ML models identify more non-EM kidneys while recommending fewer EM kidneys for OOS allocation.} Each point corresponds to one value of $\beta$ and shows low-KDPI non-EM kidneys identified for potential OOS placement (y-axis) against low-KDPI EM kidneys unnecessarily routed OOS (x-axis). Stars mark $\beta=0.5$.}
\label{fig:frontier}
\end{figure}

We assign a cost of 1 to keeping a non-EM organ on-sequence, risking discard (a false positive) and a cost of $\beta$ to unnecessarily moving an EM organ OOS, unfairly bypassing priority order (a false negative). A model's average decision cost is the mean of these accumulated costs over all test organs. For each calibrated predictor, the preference parameter $\beta$ induces a trade-off between keeping EM organs on-sequence and identifying non-EM organs for OOS allocation. Figure~\ref{fig:frontier} traces the resulting trade-off on low-KDPI test kidneys as $\beta$ varies. Curves further up and to the left are better: they identify more low-KDPI non-EM organs while recommending fewer EM organs for OOS allocation.
The ML models improve on the calibrated \KDPI{} baseline, and the trade-off generally improves with model complexity: \XGB{} generally outperforms \RF{}, which outperforms CART, which outperforms logistic regression. The Arbitrator, which is discussed further in the next section, is shown for completeness in Figure \ref{fig:frontier}, but we observe that the Arbitrator approximates \RF{} performance very closely. We use the starred operating point at $\beta = 0.5$ (which implies a decision threshold of $\tau_{0.5} = 2/3$) for the remaining analyses, which weights the benefit of saving an organ from discard at twice the benefit of maintaining the match run.
At $\beta = 0.5$, \XGB{} identifies 66.7\% of low-KDPI, non-EM kidneys (3,300 of the total 4,944) for potential OOS placement while unnecessarily routing only 13.7\%  of low-KDPI, EM kidneys (1,870 of 13,612) out-of-sequence. In contrast, the calibrated \KDPI{} baseline correctly identifies 485 \textit{fewer} low-KDPI non-EM kidneys for OOS placement than \XGB{} while unnecessarily routing 1,111 \textit{more} EM kidneys OOS, suggesting that a single clinical heuristic is too coarse for the inherent complexity of the prediction.

Standard predictive metrics show the same ordering (Appendix, Table~\ref{tab:si-full-diagnostics}): \XGB{} has the highest accuracy and F1 scores and the lowest log loss and average decision cost per organ, followed by \RF{}, CART, logistic regression, and the \KDPI{} baselines. Importantly, CART and \RF{}, the two models on which our interpretable approach builds, recover much of \XGB{}'s improvement over the calibrated clinical baseline. At $\beta=0.5$, their accuracies, 0.788 and 0.801, are within 0.027 and 0.014, respectively, of \XGB{}'s accuracy of 0.815, compared with the 0.076 gap when comparing \XGB{} with calibrated \KDPI{}. Thus, CART and \RF{} clearly outperform the clinical baseline while not deviating too far from black-box \XGB{}'s performance. However,
none of these models' point predictions convey whether the evidence behind each prediction is strong enough to act on. We next build an interpretable \textit{Arbitrator} model to restrict automation to cases with sufficient certainty.

\subsection{Interpretable Resolution with the Arbitrator}

\begin{figure}[ht]
\centering
\includegraphics[width=.7\linewidth]{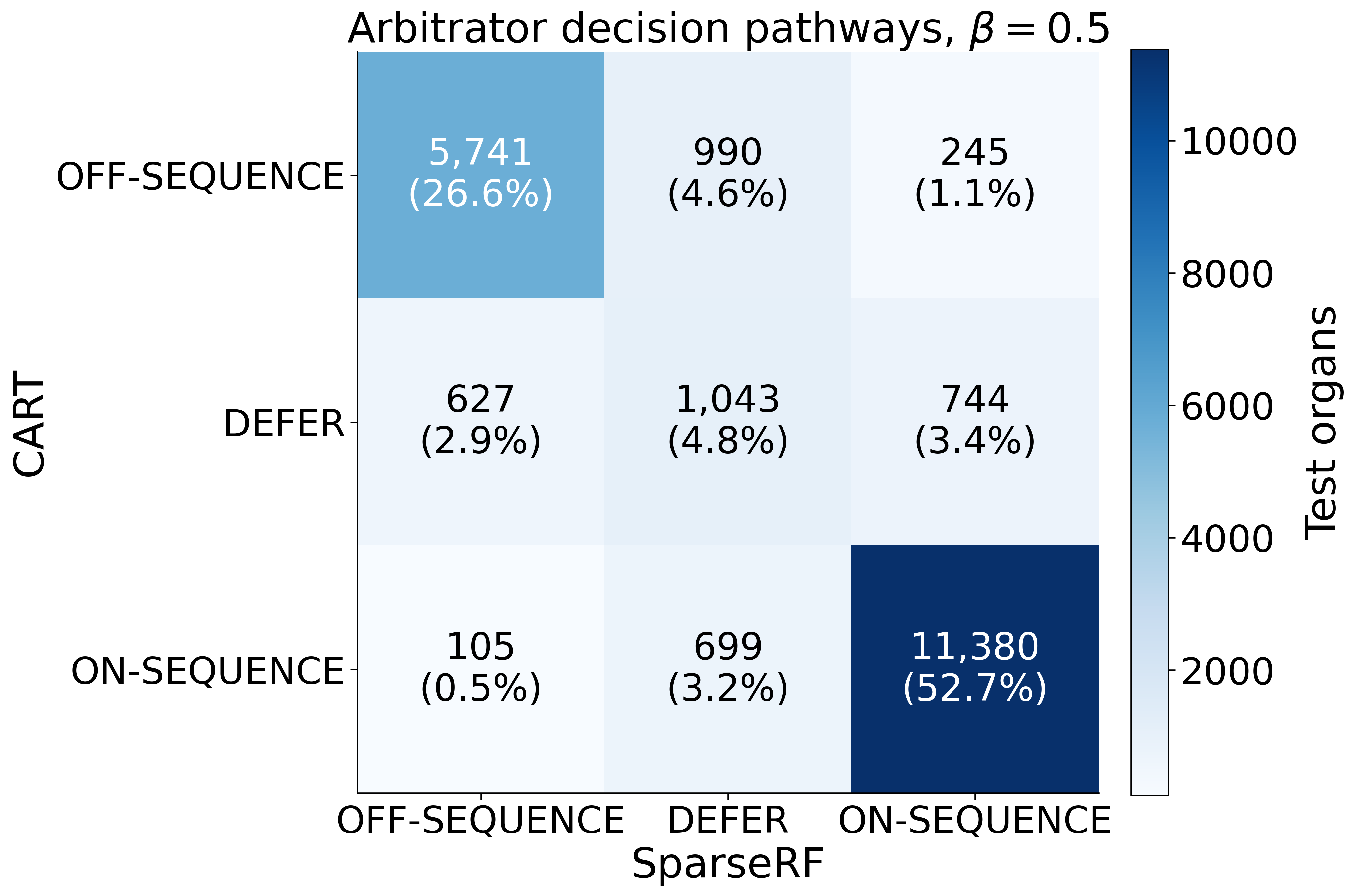}
\caption{\textbf{CART and SparseRF provide complementary certification.} Test kidney certification level at $\beta=0.5$ by CART and SparseRF.
}
\label{fig:arbitration_complementarity}
\end{figure}

We use two interpretable models: CART, a single decision tree, and SparseRF, a sparsification of the trained \RF{} that retains a small set of its rules approximating the full \RF{} model. The Arbitrator considers the optimal action prescribed by each model and the relative certainty that the prescribed action would still be optimal on out-of-sample data with high probability.
A model \textit{certifies} an action only when that action remains cost-minimizing throughout its interval; otherwise, it \textit{abstains} from making a decision.
We say an organ receives \textit{dual certification} if both models certify the same action and \textit{single certification} if only one of the two interpretable models certifies an action while the other abstains. Our Arbitrator \textit{resolves} an organ's routing decision so long as it is single- or dual-certified, and defers all other organs for individual review (see Figure \ref{fig:framework}). In this study, individual review is represented by the \XGB{} decision. Certification decisions by interpretable model on the held-out test set are shown in Figure~\ref{fig:arbitration_complementarity}.

In Figure~\ref{fig:arbitration_complementarity}, we see that for $\beta=0.5$, CART and SparseRF provided dual certification of the same action for 17,121 kidneys (79.4\% of the test set). Single certification pathways covered another 3,060 (14.2\%): 1,689 through CART alone and 1,371 through SparseRF alone. The remaining 1,393 (6.5\%) had no agreed certified action. The Arbitrator therefore resolves 93.5\% of organs with interpretable rules. The remaining 6.5\% are unresolved and always deferred for expert review. We illustrate some rules learned by the Arbitrator in Table~\ref{tab:certification-profile-examples}.
\begin{table*}[t]
\centering
\caption{\textbf{Interpretable Arbitrator rules and certification pathways at $\beta=0.5$.}
A model certifies on-sequence (OS) or out-of-sequence (OOS) placement when its entire 99\%
confidence interval lies above or below the decision threshold $\tau_\beta=2/3$,
respectively; otherwise it defers. C denotes a CART leaf and F a SparseRF rule. Additional certification profiles and selected audited rules are presented in the Appendix. \textit{Abbreviations: Cr, terminal creatinine; GS, glomerulosclerosis;
IF, interstitial fibrosis; BUN, blood urea nitrogen; HCV NAT, hepatitis C nucleic acid test.}}
\label{tab:certification-profile-examples}
\footnotesize
\setlength{\tabcolsep}{4pt}
\renewcommand{\arraystretch}{1.15}
\renewcommand{\tabularxcolumn}[1]{m{#1}}
\begin{tabularx}{\textwidth}{@{}>{\raggedright\arraybackslash}X >{\raggedright\arraybackslash}m{2.3cm} >{\centering\arraybackslash}m{2.5cm} >{\centering\arraybackslash}m{2.5cm} >{\centering\arraybackslash}m{2.4cm}@{}}
\toprule
\textbf{Donor features} & \textbf{Rule(s)}
  & \shortstack{\textbf{CART}\\\textbf{99\% CI}} & \shortstack{\textbf{SparseRF}\\\textbf{99\% CI}} & \textbf{Pathway}\\
\midrule
KDPI $>90.5\%$; GS $>20\%$; vascular changes $\ge10\%$
  & C164, F153\_400 & \shortstack{0.032--0.067\\(OOS)} & \shortstack{0.023--0.057\\(OOS)} & \shortstack{Dual-certified\\\textbf{OOS}}\\
\midrule[0.3pt]
KDPI $\le41.5\%$; Cr $\le2.595$ mg/dL; IF $<10\%$; GS $\le10\%$;
  lipase $>1290.5$ U/L; no kidney biopsy; HCV NAT negative
  & C19, F25\_13, F81\_9, F164\_278, F206\_243 & \shortstack{0.918--0.939\\(OS)} & \shortstack{0.888--0.932\\(OS)} & \shortstack{Dual-certified\\\textbf{OS}}\\
\midrule[0.9pt]
$78.5\%<$ KDPI $\le90.5\%$; GS $\le10\%$; recovery year through 2019;
  Cr $\le2.295$ mg/dL; continued smoking not recorded; BUN $>16.5$ mg/dL
  & C130 & \shortstack{0.499--0.663\\(OOS)} & \shortstack{0.000--1.000\\(defer)} & \shortstack{Single-certified\\\textbf{OOS}}\\
\midrule[0.3pt]
$27.5\%<$ KDPI $\le43.5\%$; $2.595<$ Cr $\le3.765$ mg/dL;
  sodium $>142.5$ mEq/L
  & C49, F25\_59 & \shortstack{0.557--0.786\\(defer)} & \shortstack{0.788--0.882\\(OS)} & \shortstack{Single-certified\\\textbf{OS}}\\
\midrule[0.9pt]
KDPI $\le41.5\%$; Cr $\le2.595$ mg/dL; IF $\ge10\%$
  & C21 & \shortstack{0.486--0.710\\(defer)} & \shortstack{0.000--1.000\\(defer)} & \textbf{Deferred}\\
\bottomrule
\end{tabularx}
\end{table*}
We observe that the learned decision rules generally match intuition; for example, kidneys with low values of \KDPI{}, creatinine, interstitial fibrosis and glomerulosclerosis values are more likely to be EM, whereas high values are more likely to be non-EM and recommended for OOS placement. When there is a mix of positive and negative contributing factors, single-certification pathways and deferral are more common.

\subsection{Modest Cost of Interpretability}
It is well-known that interpretability comes at a cost, but we ask where this cost arises and how large the decision cost gap is versus our black box \XGB{}. We investigate the relative costs of misclassifications on historical data in Table \ref{tab:arbitration-certification-errors-beta-0-5}, recognizing the caveat that low prediction error alone is insufficient for making future, high-quality recommendations. We find in Table \ref{tab:arbitration-certification-errors-beta-0-5} that the misclassification rates vary significantly by the Arbitrator's certification pathway. In particular, resolving decisions with interpretable rules comes at essentially no cost relative to \XGB{} on dual certified pathways; most of the relative increase in error and decision cost comes from single-certified pathways, allowing us to separate the regions of higher and lower certainty. We next discuss these high-level findings with more granularity.

\begin{table}[hbt]
\centering
\caption{\textbf{Performance by certification pathway at $\beta=0.5$.}
FPR and FNR denote false positive and false negative rates, respectively. The positive class corresponds to EM organs. For deferred kidneys, only XGBoost predictions are available. Certification pathway ``Any'' includes all 21,574 test kidneys; its Arbitrator metrics include XGBoost decisions for deferred kidneys.}
\small
\setlength{\tabcolsep}{4pt}
\renewcommand{\arraystretch}{1.12}
\begin{tabularx}{\columnwidth}{@{}Xrrrr@{}}
\toprule
\shortstack[l]{Certification\\Pathway} & Accuracy & FPR & FNR & Avg. cost \\
\midrule
\multicolumn{5}{@{}l}{\textbf{Any}} \\
\addlinespace[2pt]
\XGB{}     & 0.815 & 0.223 & 0.165 & 0.131 \\
Arbitrator & 0.802 & 0.252 & 0.170 & 0.143 \\
\midrule
\multicolumn{5}{@{}l}{\textbf{Dual certified}} \\
\addlinespace[2pt]
\XGB{}     & 0.853 & 0.208 & 0.116 & 0.108 \\
Arbitrator & 0.849 & 0.226 & 0.113 & 0.114 \\
\midrule
\multicolumn{5}{@{}l}{\textbf{CART-only certified}} \\
\addlinespace[2pt]
\XGB{}     & 0.658 & 0.246 & 0.415 & 0.224 \\
Arbitrator & 0.565 & 0.323 & 0.518 & 0.287 \\
\midrule
\multicolumn{5}{@{}l}{\textbf{SparseRF-only certified}} \\
\addlinespace[2pt]
\XGB{}     & 0.653 & 0.284 & 0.380 & 0.222 \\
Arbitrator & 0.606 & 0.407 & 0.387 & 0.267 \\
\midrule
\multicolumn{5}{@{}l}{\textbf{Deferred}} \\
\addlinespace[2pt]
\XGB{} & 0.704 & 0.300 & 0.293 & 0.202 \\
Arbitrator & --- & --- & --- & --- \\
\bottomrule
\end{tabularx}
\label{tab:arbitration-certification-errors-beta-0-5}
\end{table}

Across all test kidneys, the Arbitrator increases average decision cost by 9.1\% compared to \XGB{} while resolving 93.5\% of organs with interpretable rules. Most of the 0.014-point accuracy gap can be attributed to the 0.029-point gap in the Arbitrator's false positive rate relative to \XGB{}, meaning that the Arbitrator incorrectly keeps more non-EM kidneys on-sequence than \XGB{}. For dual-certified organs, the Arbitrator nearly matches \XGB{}'s average cost of 0.108 with only a 4.8\% increase to 0.114; for single-certified organs, the gap in the cost attained widens to 28.0\% for CART-only certification, and to 20.0\% for SparseRF-only certification. The cost gap is localized to the difficult single-certified cases: although single-certified kidneys make up only 14.2\% of the test set, they account for about two-thirds of the Arbitrator's additional cost. Note that decision-making on organs receiving single-model certification was generally harder for \textit{all} models.  In terms of potential impact, the Arbitrator retroactively recommended 1,920 of 2,738 (70.1\%) low-KDPI, \textit{historically discarded} kidneys for OOS placement compared with 2,079 (75.9\%) for \XGB{} (Appendix, Table~\ref{tab:si-potentially-saved}). That is, if our model were deployed retroactively, anywhere between 1,920 and 2,079 patients on the wait list could have received a \textit{high-quality} organ but were never given the opportunity. Thus, while we see a trade-off in predictive power between interpretable and non-interpretable models, our model demonstrates a high potential impact for saving historically discarded, low-KDPI organs.

We next discuss the 1,393 (6.5\%) kidneys that the Arbitrator deferred: 1,043 for which both CART and SparseRF abstained due to uncertainty, and 350 for which they certified opposite actions. On this deferred subset, XGBoost has lower accuracy (0.704 vs.\ 0.853) and higher average cost (0.202 vs.\ 0.108) than on dually certified kidneys.
On the deferred subset, however, \XGB{} has higher accuracy and lower average cost than on either single-certified subset.
This pattern suggests that some decisions left unresolved by the interpretable models may include decisions better captured by more complex models like XGBoost, while single-certified kidneys remain distinctly challenging, even for XGBoost. A possible hypothesis is that single-certified organs have higher amounts of inherent or \textit{aleatoric} uncertainty in their outcomes, for which additional model complexity or more data points can do little to improve predictive performance. On the other hand, the deferred organ subset may have more reducible or \textit{epistemic} uncertainty that \XGB{} is able to reduce through increased model complexity. Regardless of the merit of this hypothesis, deferred and single-certified kidneys identify different types of organs on which prediction is challenging.

\begin{figure*}[t!]
\centering
\includegraphics[width=0.95\textwidth]{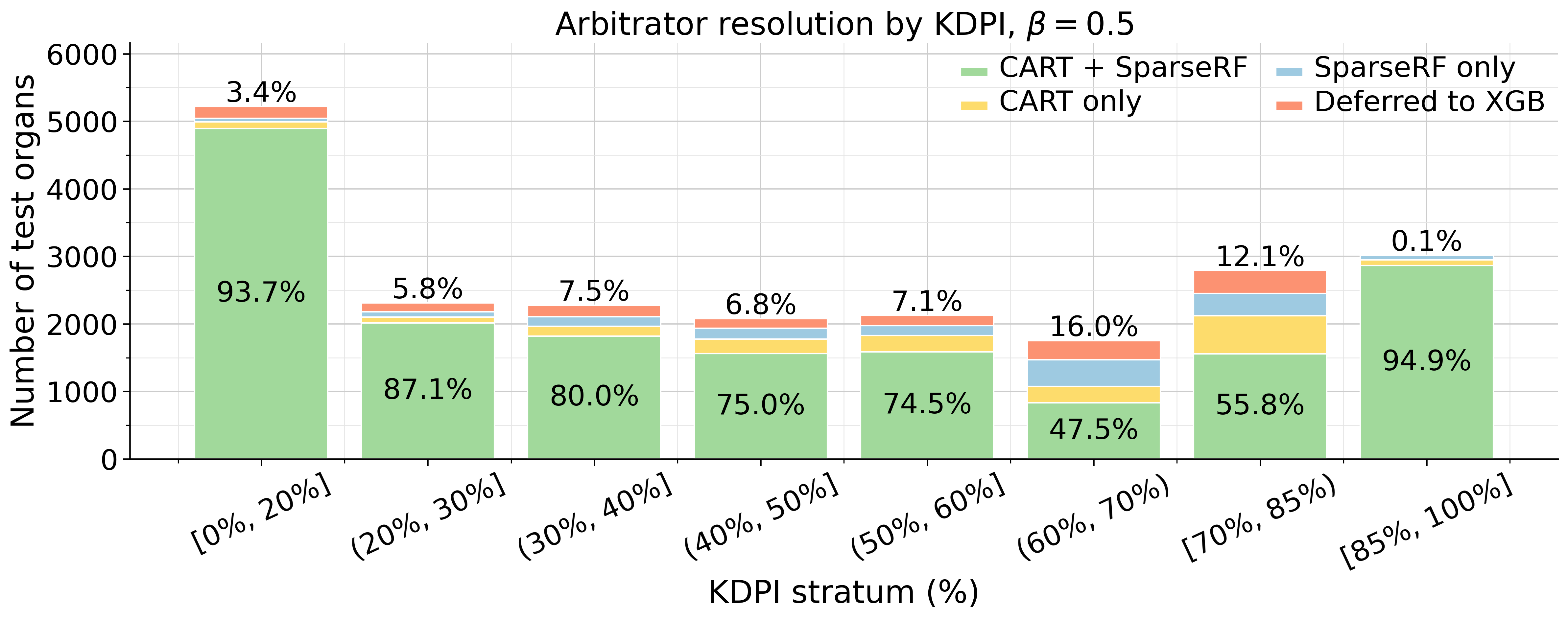}
\caption{\textbf{Arbitration pathway varies across KDPI.}
Certification pathways for 21,574 held-out kidneys at $\beta=0.5$, grouped by KDPI. }
\label{fig:arbitration}
\end{figure*}

We next examine where these harder cases arise in terms of the perceived quality of the organ. Figure~\ref{fig:arbitration} considers how deferral, single certification, and dual certification rates vary with \KDPI{}, a normalized metric of organ quality where lower values are preferred.
We observe that single certifications and deferrals are most concentrated for organs with \KDPI{} values between 60\% and 85\%. Deferral from model disagreement or uncertainty more than doubles from 7.1\% to 16.0\%, and single certification nearly doubles from 18.4\% to 36.5\% in the 60--70\% range. The dual certification rate remains low (55.8\%) in the 70--85\% range but shoots back up to 94.9\% after the clinically relevant 85\% threshold.
Thus, we observe that these difficult single-certification or uncertified cases happen at intermediate \KDPI{} values, whereas 93.7\% and 94.9\% of kidneys with KDPI at most 20\% or at least 85\%, respectively, are dually certified. This result is intuitive: for organs that are cleanly high- or low-quality, the decision is clear, and the average cost of these decisions is low. The predictions are significantly more difficult at intermediate \KDPI{} values where further deterioration can occur if matching does not take place quickly.

\subsection{Auditing the Arbitrator Pathways}
We have shown that the Arbitrator uses interpretable rules that can be checked before automation, while maintaining competitive performance compared to the black-box \XGB{} model.
However, this performance is measured against historical decisions, which should not be treated as ground truth.
Instead of burdening clinicians with auditing 21,574 individual organs, we can instead have them audit the 113 rules from single-certification and dual-certification pathways. In this work, we show a proof of concept for this approach using a large language model.

Before any decision is automated, its relevant rule must be reviewed and receive one of three outcomes: provisionally retained (decide), referred back to human decision-makers on an organ-by-organ basis (defer), or flagged for the opposite action due to new policy or research (deviate), as shown in Figure \ref{fig:framework}. Our exploratory audit assessed these 113 certified contributing rules---35 CART subtrees and 78 SparseRF rules---using literature selected by the agent and assignment criteria developed by the agent. It recommended provisionally retaining 62 actions (decide), referring 43 for physician interpretation (defer), and reconsidering eight opposite actions (deviate). All eight candidates propose on-sequence rather than learned out-of-sequence placement: six concern acute kidney injury (AKI) and two concern a sclerosis threshold. If we assume that the LLM's audit recommendations are well-founded, up to 86.6\% (17,471 of 20,181) of the originally resolved kidneys or 81.0\% of all 21,574 test kidneys remain eligible for automation. Furthermore, we find that single-certified rules are flagged more often at the audit stage than dual-certified rules. Under the hypothetical audit procedure, 32.4\% of single-certified kidneys (990 of 3,060) were recommended for additional review, compared with only 10.0\% of dually certified kidneys (1,720 of 17,121; Appendix, Table~\ref{tab:si-audit-policy-pathways}).

\begin{table}[!htbp]\centering\small
\caption{Rules audited by the LLM agent. In both rules, the recommendation from the Arbitrator is to place the organs out-of-sequence (OOS); however, the audit proposed keeping them on-sequence (OS), based on \cite{Liu2020,Wen2024,Na2026}. \textit{KDPI: Kidney Donor Profile Index; Cr: terminal creatinine (mg/dL); IF: interstitial fibrosis; GS: glomerulosclerosis}.}\label{tab:audit-examples}
\setlength{\tabcolsep}{3pt}
\renewcommand{\arraystretch}{1.12}
\begin{tabularx}{\linewidth}{@{}l>{\raggedright\arraybackslash}Xc@{}}\toprule
Rule & Condensed description & Conf. Interval \\\midrule
C36 & $41.5\%<\mathrm{KDPI}\le60.5\%$; $1.675<\mathrm{Cr}\le2.595$; IF $<10\%$; GS $\le10\%$; no coronary angiogram; recovery after 2019
& $[.377,.645]$ \\
C51 & KDPI $\le43.5\%$; Cr $>3.765$; cause of death not recorded as anoxia
& $[.158,.534]$\\
\bottomrule
\end{tabularx}
\end{table}

Though a more extensive clinical review is needed to verify the utility and limitations of the audit, we find interesting results from the audit already. Two of the rules the audit highlights for potential deviation on the basis of AKI are shown in Table \ref{tab:audit-examples}. For example, CART leaf C51 recommends OOS placement when KDPI is at most 43.5\% or unrecorded, terminal creatinine exceeds 3.765 mg/dL, and cause of death is not recorded as anoxia. Our audit proposes reversing the recommendation and considering OS placement instead. The audit cites recent studies of selected transplanted kidneys that suggest that AKI---indicated by high levels of creatinine---need not entail worse graft survival without additional contributing factors ~\citep{Liu2020,Wen2024,Na2026}. In the preceding example rule, KDPI is quite low, and the absence of anoxia similarly contributes positively toward transplantability, so the recent research indicates that perhaps the historic rejection behavior may no longer be justified, and, at the very least, makes a case for reviewing current practices in the transplantation system. The Appendix (\S~\ref{sec:si-audit-sample}, Table~\ref{tab:si-audit-examples}) presents all eight candidates and comparison profiles.

\section{Discussion}

While black-box ML models can accurately predict historical outcomes in complex datasets, high-stakes decision-making cannot rest on merely replicating historical practice. When these decisions are both time-sensitive and life-critical, automating the ``easy'' cases frees a limited number of human experts to focus on the truly challenging cases.
Our contribution is an Arbitrator policy that certifies an interpretable decision rule for possible automation only when (a) the recommendation is sufficiently stable over its relative uncertainty and (b) the rule is audited against clinical literature and explicitly approved for automation.

Deceased-donor kidney allocation is such a high-stakes setting: kidneys deteriorate as the match run proceeds, so OPOs must decide relatively quickly whether to continue on-sequence. Automating routing decisions for deceased-donor organs has the potential to reduce organ discard rates while simultaneously ensuring that the fair, standardized offer sequence is used as much as possible.
The Arbitrator resolves 93.5\% of organs with interpretable rules and is able to flag 70.1\% of historically discarded low-KDPI test kidneys for expedited placement (Appendix, Table~\ref{tab:si-potentially-saved}).
If this test-set performance generalizes, the Arbitrator would flag 43.9\% of historically discarded kidneys for potential expedited placement. Applying that share to the 8,418 deceased-donor kidneys recovered for transplant but not used in 2023~\citep{israni2025} gives a pool of approximately 3,700 kidneys that could have been targeted for expedited placement and potentially saved.

\paragraph{Future Work}
It is well-known that interpretability comes at some cost to predictive performance, and this is true in our setup as well. We believe this cost is worth paying: self-auditing pipelines let us inspect the trends an AI model has learned, which matters for the ethical deployment of automated systems. We leave the exploration of generation of organ-specific confidence intervals using more complex models like \XGB\ or large language models to future work. 
While our agentic auditor design is intended to be a proof of concept for how LLMs can parse interpretable rules against clinical literature, we leave development of more powerful pipelines to future work.

In the organ allocation context, we use the lack of a candidate's acceptance by original PTR sequence 100 as a proxy for difficult placement, supported by historical acceptance data (Appendix, \S~\ref{sec:si-cutoff} and Fig.~\ref{fig:offer_level_stats}). Though this is not a clinical metric, other ML work for predicting kidney acceptance outcomes has truncated the PTR sequence to denoise outcomes \citep{bertsimas2017}.
Further, our model predicts the likelihood of organ placement rather than the likelihood of a successful transplant. Since we consider the OS versus OOS allocation decision rather than the success of candidate/donor matching, we believe that predicting the likelihood of organ acceptance is more relevant than predicting post-transplant outcomes; however, transplantation outcomes data would be useful in both (re)designing the organ matching process and refining the auditor in future work.
Finally, our current framework determines \emph{whether} a kidney should leave the standard sequence, not \emph{where} it should go. This requires the development of a carefully constructed routing system that can account for donor/patient compatibility, ischemia time, and fairness across centers and patients. We leave this to future work.

Our framework decides where models are jointly confident, and defers or deviates where their interpretations disagree or are uncertain. More work is needed on the ``human'' side of this proposed human--AI integration: how uncertainty, deferral, and deviation interact with domain experts, and whether they build trust in the system. Studies of how clinicians and OPO staff interpret, trust, or override certified rules could inform the pre-review audit and the elicitation of our relative preference between sending a non-EM organ on-sequence versus sending an EM organ out-of-sequence.
In organ allocation, it also remains open how to separate correlative rules learned by the models (e.g., donor height below 65 cm) from causal rules that withstand clinical scrutiny (e.g., donor age below two years). Refitting or leave-one-out analyses on subpopulations covered by questionable rules could separate clinically meaningful splits from correlated ones. Language-model-assisted review could scale this auditing by retrieving literature, flagging rules for clinical review, and proposing revisions with their estimated performance cost. We believe that our work opens significant avenues for future work, which could help build auditable and trustworthy AI-driven pipelines, specifically for life-critical domains like healthcare.

\section*{Author contributions}
{M.P. and T.D. developed the codebase for this work. M.P. and S.G. developed the Arbitrator model, incorporating feedback and discussions with all authors. All authors contributed to the research design, analysis and interpretation of results, and writing of the manuscript.}

\section*{Competing interests}
Author Declaration: The authors have no conflicts of interest to disclose.

\section*{Acknowledgments}

We would like to acknowledge generous support from MIT-MGB HEALS initiative, and a seed grant from SERC MIT. This work was also supported by NIH grant 5R01DK131543-03 (H.Y.) and in part by the NSF Engineering Research Center for Advanced Technologies for the Preservation of Biological Systems (ATP-Bio grant EEC 1941543) (M.T. and K.U.). We would also like to acknowledge Nikolaos Trichakis and Georgia Perakis for their support through the project; preliminary data exploration by Mehrdad Ghadiri, Raghav Manohar Jayanthi, and Ju Young Lee; and a description and categorization of donor variables for preliminary ML models by David Cron.

\clearpage
\bibliographystyle{unsrtnat}
\bibliography{references}
\clearpage
\appendix
\captionsetup{font=footnotesize}
\captionsetup[subfigure]{font=footnotesize}
\captionsetup[subtable]{font=footnotesize}
\setcounter{section}{0}
\setcounter{figure}{0}
\setcounter{table}{0}
\setcounter{equation}{0}
\renewcommand{\thesection}{S\arabic{section}}
\renewcommand{\thefigure}{S\arabic{figure}}
\renewcommand{\thetable}{S\arabic{table}}
\renewcommand{\theequation}{S\arabic{equation}}
\renewcommand{\theHsection}{appendix.\arabic{section}}
\renewcommand{\theHfigure}{appendix.\arabic{figure}}
\renewcommand{\theHtable}{appendix.\arabic{table}}
\renewcommand{\theHequation}{appendix.\arabic{equation}}
\section*{Appendix}
\section{Data, cohort, and outcome}\label{sec:si-cohort}

\subsection{Data sources and the match run}\label{sec:si-matchrun}

The study combines Organ Procurement and Transplantation Network (OPTN) Potential Transplant Recipient (PTR) match-run records with Standard Transplant Analysis and Research (STAR) deceased-donor and waitlist registry data. PTR records preserve the original ranked offer sequence, offer responses, and recorded refusal reasons, whereas the deceased-donor registry supplies donor characteristics and kidney-specific measurements and dispositions. Donor identifiers link the offer histories to registry-derived kidney and donor features.

Transplant programs evaluate candidates and register them on the national waiting list. When a deceased-donor organ becomes available, the matching system combines donor and candidate information to generate a \textit{match run}: a ranked list of biologically compatible candidates under OPTN allocation policy. Kidney matching considers medical and logistical factors, including blood type, immune compatibility, waiting time, expected post-transplant survival (EPTS) scores, and geography~\cite{unos_how_we_match}. The organ procurement organization (OPO) communicates offers to transplant programs, which review donor information and may decline or express provisional interest. A provisional acceptance indicates interest in accepting the organ or obtaining further information; it is not final acceptance. When a candidate is the next in the PTR sequence, the program must officially accept or refuse the offer; if it declines, offering can continue to the next candidate in match-run order~\cite{optn_policies}. Prespecified acceptance criteria and offer filters can bypass offers (i.e., automatically decline them) without a new clinical review of each offer~\cite{unos_offer_filters_2023}. PTR histories therefore include bypasses, refusals, provisional interest, and final acceptance: an original sequence number represents a ranked position rather than a number of deliberate rejections (i.e., some of the preceding offers may have been automatically bypassed).

For model development and evaluation, we pool match-run years 2016--2019 and 2023--2025, with observations through June 30, 2025, excluding COVID-19-disrupted years 2020--2022\footnote{During the initial U.S. COVID-19 surge, national registry studies documented fewer kidney waitlist registrations and deceased-donor kidney transplants than expected, with larger reductions in states with higher COVID-19 burden~\cite{boyarsky2020covid}. Kidney recovery, utilization and discard patterns also changed during 2020~\cite{li2021covid}. These findings motivate our design choice to remove the pandemic period from consideration in our study.}. These disjoint periods span changes in allocation practice and recording conventions, so we maintain the organ recovery year as a feature in the ML models\footnote{Match-run year determines source inclusion. Of the 71,913 eligible donors, 28 have recovery dates in adjacent years: two in 2015 and 26 in 2020, with none in 2022. Their 56 kidneys remain in the pooled cohort.}.
We separately reconstruct the COVID-19-disrupted 2020--2022 match-run cohort using the same source-linkage, offer-filtering, kidney-eligibility, and outcome rules to show how OOS allocation and acceptance outcomes have shifted over time without adding these kidneys to our analysis.

\subsection{Early-matchable outcome}\label{sec:si-outcome}

Define \textit{early-matchability}, denoted EM, by $Y_i=+1$ only when an eligible final acceptance (denoted by \texttt{OFFER\_ACCEPT=Y}) occurs at \emph{original} \texttt{PTR\_}\allowbreak\texttt{SEQUENCE\_}\allowbreak\texttt{NUM}\,$\le100$ for kidney $i$. Provisional acceptance (denoted by \texttt{OFFER\_ACCEPT=Z}) does \textit{not} establish the outcome. A kidney without qualifying early final acceptance has $Y_i=-1$, denoted non-EM. We justify our use of the EM cut-off at \texttt{PTR\_}\allowbreak\texttt{SEQUENCE\_}\allowbreak\texttt{NUM}\,$\le100$ in \S~\ref{sec:si-cutoff}.

PTR acceptance records identify the donor but do not identify whether an offer concerns the left or right kidney. We therefore infer kidney-level EM labels from eligible final \texttt{OFFER\_ACCEPT=Y} records at original sequence 100 or earlier and the registry's kidney-specific transplantation status. For a donor with one recorded kidney, at least one early \texttt{Y} labels that kidney EM. For a donor with two recorded kidneys, no early \texttt{Y} labels neither kidney EM, whereas two or more early \texttt{Y}s label both as EM. We have to infer in the case where there is exactly one early \texttt{Y}. In this case, each kidney recorded as transplanted is labeled EM, and each kidney not recorded as transplanted is labeled non-EM. Thus, if both kidneys were transplanted, both are labeled EM despite only one observed early acceptance; if only one was transplanted, only that kidney is labeled EM. This is an assignment convention, rather than a ground truth. This single-\texttt{Y} assignment applies to 16,387 of the 71,913 donors in the primary union (22.79\%).

\subsection{Cohort construction}\label{sec:si-filtering}

\begin{figure}[!htb]\centering
\includegraphics[width=.95\linewidth]{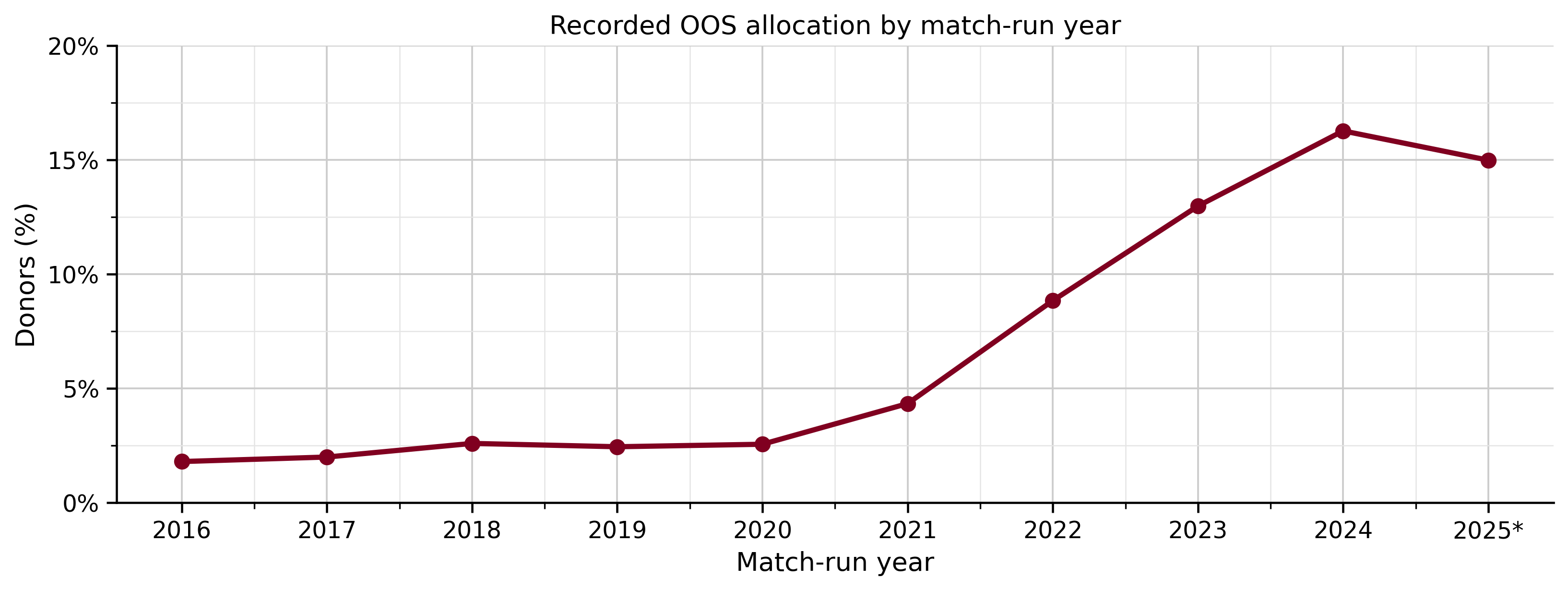}
\caption{\textbf{Recorded OOS indicators by match-run year.} The numerator is the number of distinct donors with code 861, 862, 863 or 887 in either PTR refusal field during that calendar year; the denominator is all distinct donors with kidney PTR match-run records that year, before study exclusions. The asterisk marks partial 2025 data, January 1--June 30. }\label{fig:si-oos-year}
\end{figure}

We construct the cohort by linking deceased donor records to their generated match run based on the \texttt{DONOR\_ID}. A donor can donate up to two organs, and prediction occurs at the kidney level. Each kidney receives the EM label and organ-assignment convention defined in \S\ref{sec:si-outcome}.
To study acceptance under standard match-run ordering, we exclude kidneys from donors with recorded indicators of out-of-sequence (OOS) placement. The study uses codes 861 (operational--OPO), 862 (donor medical urgency), 863 (offer not made due to an expedited-placement attempt), and 887 (not offered--expedited placement), which OPTN identifies as indicators of OOS allocation~\cite{hrsa_allocation_out_of_sequence}. A code in either refusal field triggers exclusion of the donor's kidneys. Figure~\ref{fig:si-oos-year} shows the annual fraction of PTR donors with recorded OOS indicators between 2016 and 2025. The PTR refusal-code system changed on December 2, 2021~\cite{optn_refusal_codes_2022}. We therefore apply the historical operational-refusal code list to offers dated before that change and the revised list to offers dated on or after it. This removes 248,862 and 195,406 offer records, respectively.

We first standardize donor identifiers, original PTR sequence numbers, and offer responses. We remove all offers for donors flagged \texttt{Y} for dual-kidney allocation (a missing flag counts as unflagged), then apply donor-wide OOS exclusions. The study's operational-refusal exclusions cover atypical candidate circumstances leading to rejection (i.e., temporary medical unsuitability, or a requirement for multiple organs) and logistical barriers (e.g., surgeon availability or excess travel times)~\cite{optn_refusal_codes_2022}. We remove these records based on the primary refusal field. These exclusions do not renumber the original sequence. Operational refusals use the vocabulary applicable to each match date: before December 2, 2021, codes 800, 801, 802, 803, 820, 822, 823, 825, 860; thereafter, 720, 721, 722, 723, 724, 727, 728, 760, 762, 763, 765.

Table~\ref{tab:si-preprocessing-flow} follows the kidney observations in the linked study data through eligibility screening. Of the 16,114 kidneys excluded before disposition screening, 16,056 belong to donors with observable OOS pathways and 58 have donor offer records in the source data but none remain after offer-eligibility filtering. Without an eligible offer history, these kidneys cannot be assigned the study outcome. The labeling procedure also excludes donors whose recorded final acceptances are all removed by offer filtering, rather than labeling their kidneys non-EM on that basis; this condition removes no additional kidneys outside the two groups above. Finally, 26 kidneys have either an unknown disposition or a disposition outside the retained categories 3, 4, 5, and 6.\footnote{These codes correspond to the following. 3: not recovered; 4: recovered for purposes other than transplantation; 5: recovered for transplantation but not transplanted; 6: transplanted~\cite{optn_logistics_2021}. The 26 exclusions comprise 13 kidneys with missing disposition, 12 with code 2 (authorization not obtained), and 1 with code 1 (authorization not requested).}

\begin{table}[!htb]\centering\small
\caption{Construction of the eligible pooled kidney dataset from the linked study inputs. Each row removes kidneys from the preceding row; offer-record removal counts are reported separately in the text.}\label{tab:si-preprocessing-flow}
\begin{tabularx}{\linewidth}{Xrr}\toprule
Stage & Kidneys removed & Kidneys remaining\\\midrule
Kidney observations in linked study inputs & --- & 159,961\\
Donor has an observable OOS placement pathway & 16,056 & 143,905\\
All donor offers removed by eligibility filtering & 58 & 143,847\\
Disposition outside retained categories or unknown & 26 & 143,821\\\midrule
Eligible prediction cohort & --- & 143,821\\\bottomrule
\end{tabularx}
\end{table}

\subsection{Donor-grouped splits}

We randomly assign donors, rather than individual kidneys, to training, validation, and test sets. Keeping both kidneys from the same donor together prevents their shared clinical characteristics and offer history from appearing on both sides of model development and evaluation. Before splitting, each donor is assigned to a stratum defined by the combination of KDPI decile and the number of EM kidneys from that donor (0, 1, or 2). Stratification preserves approximately similar distributions of donor risk and early acceptance across sets (Table~\ref{tab:si-cohort}).
The split has two stages. First, 70\% of donors are assigned to training and the remaining 30\% are held aside. Second, those held-aside donors are divided equally between validation and test, producing approximately 70/15/15\% donor allocation overall. Both stages use the same strata. The kidney percentages can differ slightly because donors contribute different numbers of retained kidneys. All models use the same donor assignments. Test data are reserved for final evaluation. Table~\ref{tab:si-cohort} reports the balance of the stratified split. The eligible cohort contains 143,821 organs belonging to 71,913 donors.

\begin{table}[!htb]\centering\small
\caption{Kidney counts, EM prevalence, and acceptance at any sequence in the donor-grouped splits. Kidney percentages use the full 143,821-kidney cohort; EM and Accepted percentages use the kidneys within each row. Accepted means organ-assigned eligible final acceptance at any original sequence, so EM is a subset of Accepted.}\label{tab:si-cohort}
\setlength{\tabcolsep}{12pt}
\begin{adjustbox}{max width=\linewidth}
\begin{tabular}{lrrrrrr}\toprule
Split & \shortstack{Kidneys\\(\% of cohort)} & EM & Non-EM & Accepted & \shortstack{\% EM\\in split} & \shortstack{\% Accepted\\in split}\\\midrule
Training & 100,674 (70.00\%) & 65,915 & 34,759 & 73,007 & 65.47 & 72.52\\
Validation & 21,573 (15.00\%) & 14,122 & 7,451 & 15,610 & 65.46 & 72.36\\
Test & 21,574 (15.00\%) & 14,127 & 7,447 & 15,592 & 65.48 & 72.27\\\midrule
All splits & 143,821 (100.00\%) & 94,164 & 49,657 & 104,209 & 65.47 & 72.46\\\bottomrule
\end{tabular}
\end{adjustbox}
\end{table}

\subsection{Operational populations and the 100-offer cutoff}\label{sec:si-cutoff}
We use KDPI $<85\%$ as an operational policy proxy for whether an organ should be accepted by someone on the match run if there is no further deterioration. Kidneys meeting this criterion are called \textit{low-KDPI}. We note that kidneys with KDPI above 85\% \textit{can} be transplanted; however OPTN Policy 5.3.C requires transplant programs to obtain written, informed consent before offering such kidneys to a candidate~\cite{optn_policies} due to increased risk of graft failure.

From this low-KDPI subset, we consider two base populations. The first is the set of low-KDPI, non-EM organs in the dataset. We interpret this subset as the set of high-quality, historical donor organs that demonstrated elevated risk of discard by maintaining the standard match run. While they may have eventually been accepted after PTR sequence number 100 and transplanted, we identify these organs as possibly being a ``close call'' that should be avoided. Second, we consider low-KDPI kidneys with recorded explicit discard, identified by codes 503, 504, or 531 in the registry-derived \texttt{REASON\_CD} field. Thus, we can proxy the number of transplantable organs saved by our methodology by the number of these low-KDPI originally-discarded organs that would have been recommended for OOS allocation. The 21,574-kidney test set contains 4,944 low-KDPI non-EM kidneys and 2,738 low-KDPI historically discarded kidneys, the populations used in the main-text intervention summaries. Across all training, validation, and test observations from the pooled 2016--2019 and 2023--2025 match-run years, the 143,821-kidney cohort contains 33,068 low-KDPI non-EM kidneys and 18,302 low-KDPI historically discarded kidneys.

\begin{figure}[htbp]\centering
\includegraphics[width=.45\linewidth]{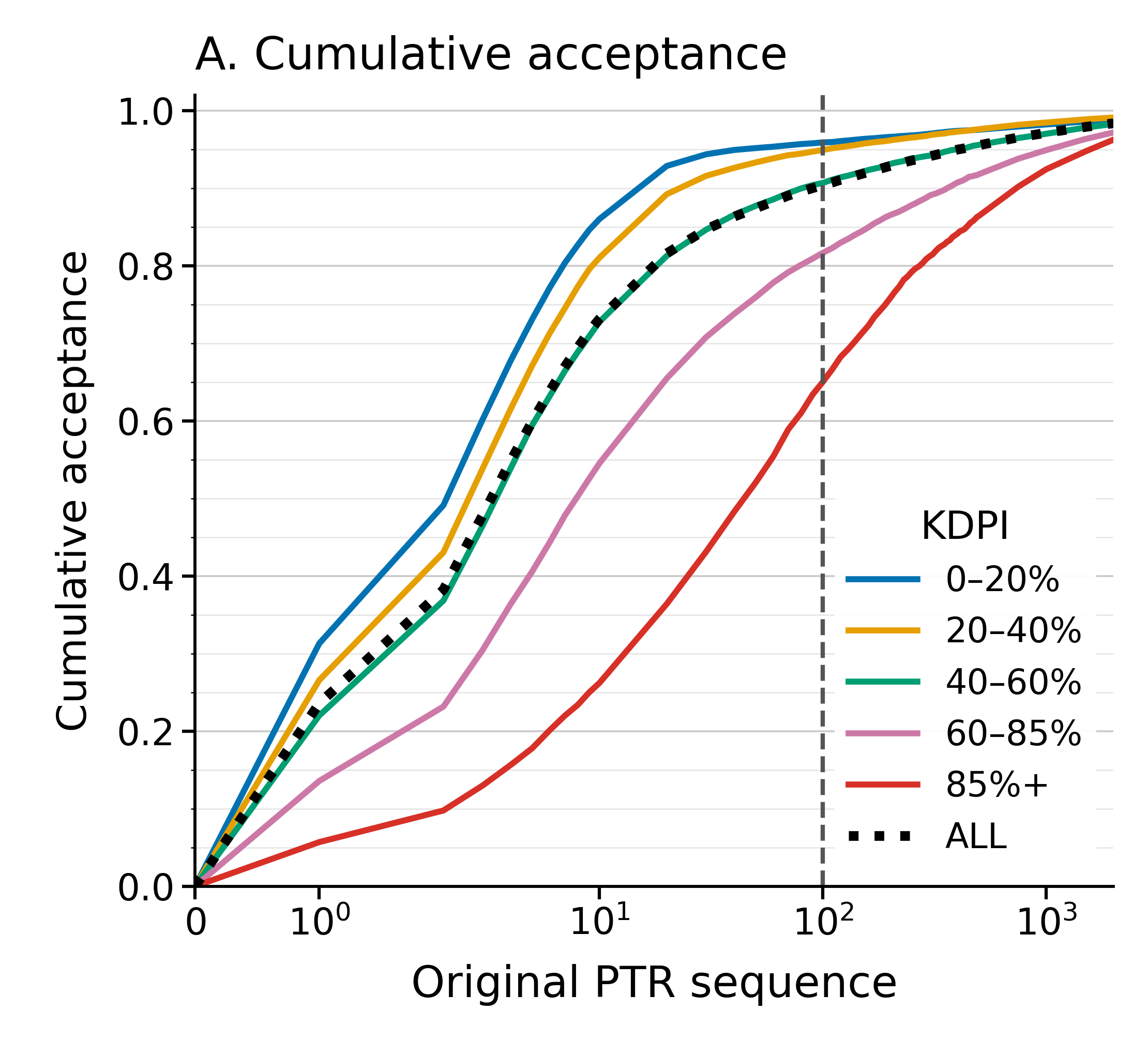}\hfill
\includegraphics[width=.45\linewidth]{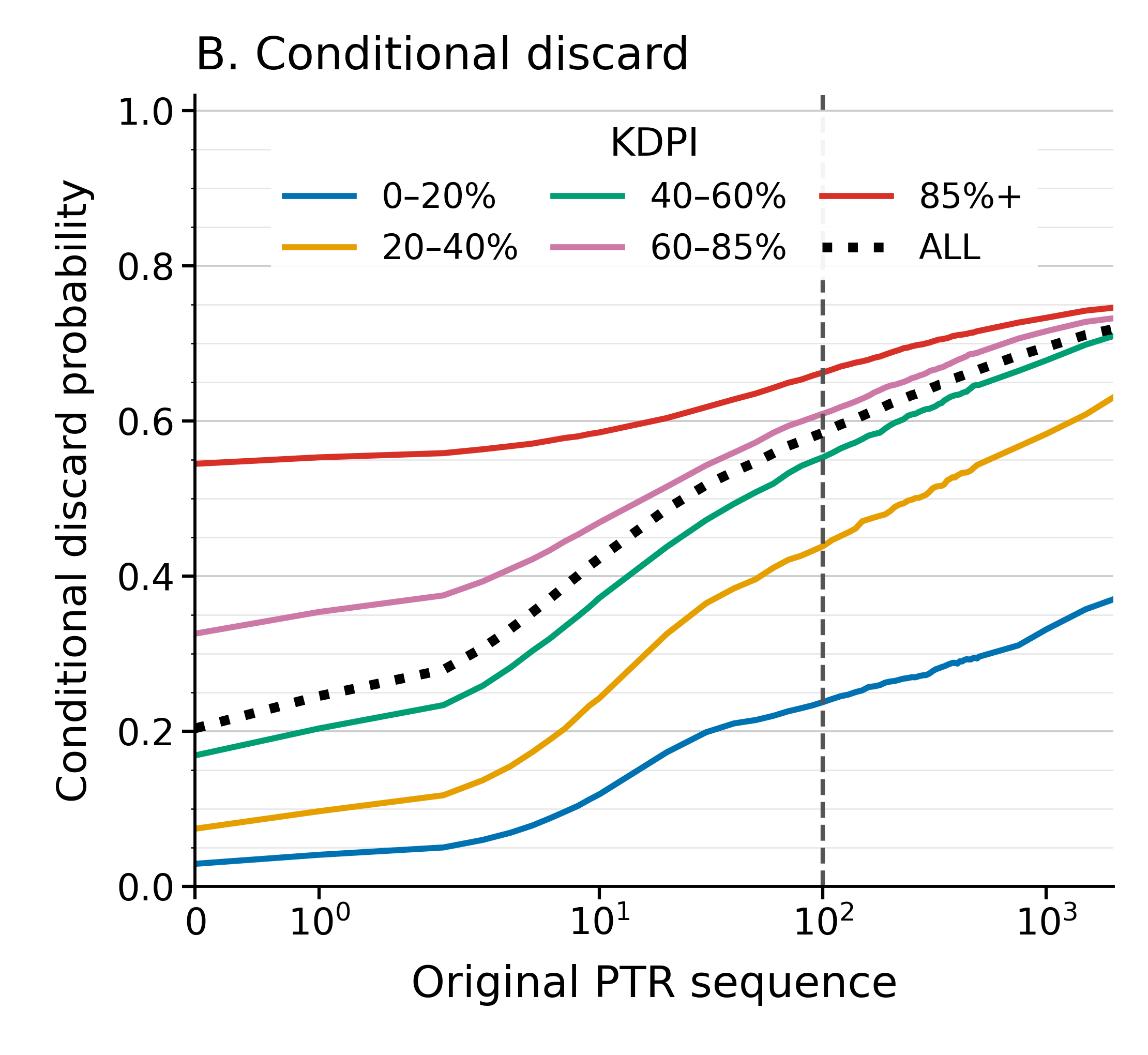}
\caption{\textbf{Empirical offer-level acceptance and discard statistics.} (A) Cumulative eligible final acceptance among eventually accepted kidneys by original PTR sequence. (B) Empirical probability of explicit organ discard among kidneys not accepted by sequence $k$. Both panels use identical KDPI bands: 0--20\%, 20--40\%, 40--60\%, 60--85\%, and 85--100\% (labeled 85\%+). Bands include their lower endpoint and exclude their upper endpoint.}\label{fig:offer_level_stats}
\end{figure}

We prespecified acceptance by original sequence 100 as the threshold for early matchability. A useful threshold should satisfy two conditions: most kidneys that are eventually accepted should be accepted before it, and kidneys still unplaced after it should face substantially higher risk of discard. Figure~\ref{fig:offer_level_stats} examines both conditions using two descriptive curves, separated by KDPI value. Cumulative acceptance at sequence $k$ is the fraction of eventually accepted kidneys that were accepted by sequence $k$. Conditional explicit discard at sequence $k$ is the fraction of kidneys still unplaced when their donor's match run reaches sequence $k$ that were ultimately discarded.

Both conditions hold at sequence 100. By that point, 91.76\% of eventually accepted kidneys with KDPI $<85\%$ had been accepted (90.36\% across all KDPI values), so the threshold captures most acceptances. Among kidneys still unplaced by the hundredth candidate on the PTR sequence, the pooled probability of explicit discard was 58.46\%, compared with 20.35\% at the start of the match run, a 2.87-fold increase. A kidney not placed by sequence 100 is therefore not necessarily unusable or destined for discard, but it does face markedly higher risk. These patterns support the 100-offer threshold as a proxy for EM.

\subsection{Baseline outcomes and year effects}
\label{sec:si-baselines}
In addition to the aggregated early acceptance, and ever-accepted statistics shown in Table~\ref{tab:si-cohort}, Table~\ref{tab:si-year-baselines} reports how these metrics vary year-over-year, including, for the sake of completeness, the pandemic years.
\begin{table}[!htb]\centering\small
\caption{Retained kidney acceptance rates by donor recovery year or epoch. Accepted denotes organ-assigned eligible final acceptance at any original sequence; EM denotes acceptance by original sequence 100. Asterisks identify rows from the excluded 2020--2022 match-run cohort. Cohorts contain minimal, but positive overlap, so their totals must not be pooled.}\label{tab:si-year-baselines}
\begin{tabular*}{.75\linewidth}{@{\extracolsep{\fill}}lrrr}\toprule
Year or epoch & Kidneys & Accepted (\%) & EM (\%)\\\midrule
2016 & 18,420 & 77.78 & 70.55\\
2017 & 18,988 & 78.74 & 71.79\\
2018 & 19,588 & 79.85 & 72.45\\
2019 & 21,874 & 79.93 & 72.36\\
2020* & 23,322 & 78.03 & 70.65\\
2021* & 25,252 & 73.06 & 64.98\\
2022* & 25,701 & 69.00 & 62.14\\
2023 & 26,213 & 66.02 & 59.36\\
2024 & 25,684 & 62.55 & 55.84\\
2025 & 12,998 & 64.61 & 58.36\\\midrule
2016--2019 & 78,870 & 79.12 & 71.82\\
2020--2022* & 74,275 & 73.21 & 65.78\\
2023--2025 & 64,895 & 64.36 & 57.77\\\midrule
COVID-excluded dataset & 143,821 & 72.46 & 65.47\\\bottomrule
\end{tabular*}
\end{table}
In Table~\ref{tab:si-year-baselines}, both final acceptance and early acceptance rates decline across the three recovery-year epochs: 79.12\% in 2016--2019, 73.21\% in 2020--2022, and 64.36\% in 2023--2025. EM prevalence likewise falls from 71.82\% to 65.78\% to 57.77\%. Within the descriptive pandemic cohort, EM prevalence declines from 70.65\% in 2020 to 62.14\% in 2022, so the contrast between the primary study's two eras spans a decline already visible during the omitted years. Across the first and last epochs, acceptance falls by 14.76 percentage points and EM prevalence by 14.05 percentage points.

Published studies provide context for this drift. Li and colleagues documented disruption of kidney recovery and utilization during the initial COVID-19 surge, with changing discard patterns across 2020~\cite{li2021covid}. Following the March 2021 introduction of circle-based allocation (KAS250), Cron and colleagues found increased offer volume and more transplant centers approached before acceptance~\cite{cron2023offers}. Bradbrook and colleagues attributed much of the recent national increase in kidney nonuse to changes in the donor pool~\cite{bradbrook2025}. These studies concern different populations and outcomes and do not quantify the causes of our EM-rate difference. They motivate retaining recovery year as a predictor; the pandemic years remain a descriptive comparison outside the primary non-COVID union cohort.

\section{Predictive models}\label{sec:si-models}

\subsection{Predictors and missingness}\label{sec:si-features}

We train six models: the raw KDPI cutoff at 85\%, calibrated KDPI cutoff, logistic regression, CART, Random Forest, and XGBoost. The first three models make use of a limited feature space for prediction. The two KDPI methods only use KDPI thresholding to make predictions. The raw-cutoff model predicts EM whenever KDPI $\leq 85\%$ and non-EM otherwise. The calibrated KDPI model fits isotonic regression to training outcomes using $1-\mathrm{KDPI}/100$ as the predictor, with KDPI expressed as a percentage. Logistic regression uses the five predictors: serum creatinine levels, age, history of diabetes, KDPI, and an indicator of whether or not the patient is a pediatric candidate.

CART, Random Forest, and XGBoost are able to make use of the full preprocessed donor feature set. We curate this feature set from 188 source inputs before categorical expansion. Donor identifiers, final disposition, transplantation, acceptance and downstream offer-count evidence are excluded from prediction; transplantation flags above are used only to construct labels. Continuous tree inputs retain native missing values. Categorical predictors are represented by indicator variables for categories observed in the training data; the same encoding is applied to validation and test data.

\subsection{Training and calibration}\label{sec:si-training}

\begin{figure}[htbp]\centering
\includegraphics[width=.8\linewidth]{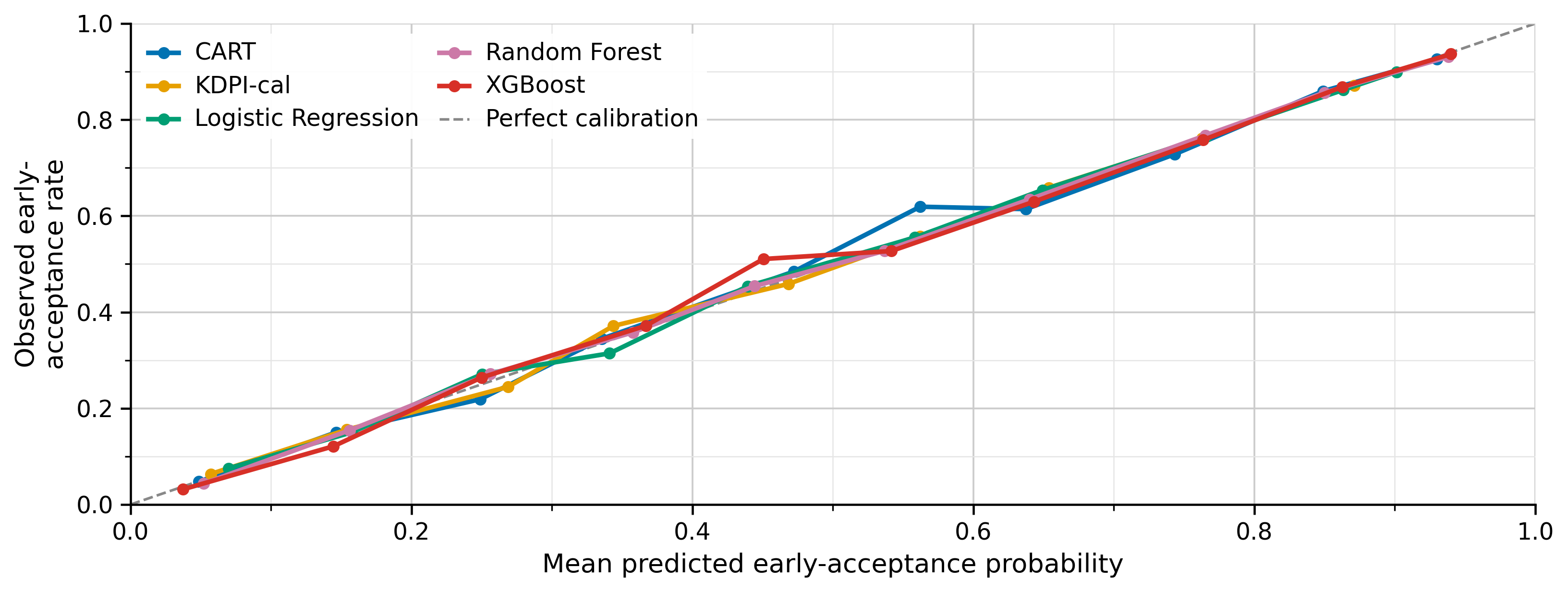}
\caption{Test-set calibration of the five probability models (21,574 kidneys). Kidneys are grouped into ten equal-width bins of predicted early-acceptance probability; each point plots the bin's mean prediction against its observed early-acceptance rate.}\label{fig:si-calibration}
\end{figure}

We use training data for model fitting and calibration, while validation data supports rule uncertainty estimates. We reserve test data for final evaluation.
In all of the ML models (i.e., excluding both KDPI benchmarks), the trained model is calibrated using isotonic regression on donor-grouped out-of-fold training predictions. For calibrated KDPI, we fit isotonic regression directly to the fixed scores $1-\mathrm{KDPI}$ and observed EM outcomes in the training set. Unlike the ML models, this score does not require fitting a predictive model, so out-of-fold predictions are unnecessary.
Every calibrated model is thresholded at the cost-minimizing value $\tau_\beta=1/(1+\beta)$, so no decision threshold is tuned on validation data. Held-out evaluation on the test set reports accuracy, log loss, F1 for each class, and average decision cost incurred from false positives (cost of 1) and false negatives (cost of $\beta$).
We see in Figure \ref{fig:si-calibration} that isotonic calibration fitted on cross-fitted training predictions transfers to the held-out test set. The kidney-weighted mean absolute gap between predicted and observed rates is 0.005--0.011 across models, and no bin with at least 200 kidneys differs by more than 0.06 (CART and XGBoost) or 0.03 (calibrated KDPI, logistic regression and Random Forest).

\subsection{Operating points and performance}
Every calibrated model uses the policy threshold $\tau_{0.5}=2/3$, the cost-minimizing threshold at $\beta=0.5$. Log loss is determined by probability scores, so the decision threshold $\tau_\beta$ is irrelevant, whereas accuracy, F1 and average cost depend on $\tau_\beta$. The Arbitrator is a decision rule built from ML models, not an additional trained probability model. Table~\ref{tab:si-full-diagnostics} reports each model's held-out performance at this operating point. All models generally rank in the same order by each metric: XGBoost, Random Forest, CART, logistic regression, then the KDPI baselines.

\begin{table}[!htb]\centering
\caption{Individual-model performance on the same held-out test set ($N=21{,}574$ kidneys) at the fixed $\beta=0.5$ policy threshold $\tau_{0.5}=1/(1+\beta)=2/3$ for every calibrated model. The raw KDPI benchmark uses its fixed 85\% cutoff. Average cost is $(\mathrm{FP}+0.5\,\mathrm{FN})/N$. F1$+$ and F1$-$ computes the F1 metric for the positive (EM) and negative (non-EM) class, respectively.}\label{tab:si-full-diagnostics}
\footnotesize\setlength{\tabcolsep}{6pt}
\begin{adjustbox}{max width=\linewidth}
\begin{tabular}{lrrrrr}\toprule
Model & Accuracy & Log loss & F1$+$ & F1$-$ & Avg. cost\\\midrule
KDPI cutoff & .744 & .569 & .832 & .463 & .245\\
Calibrated KDPI & .739 & .499 & .791 & .654 & .180\\
Logistic regression & .758 & .477 & .806 & .680 & .165\\
CART & .788 & .445 & .835 & .707 & .151\\
Random Forest & .801 & .423 & .844 & .726 & .140\\
XGBoost & .815 & .391 & .856 & .744 & .131\\\bottomrule
\end{tabular}
\end{adjustbox}
\end{table}

\section{Validation-supported uncertainty and SparseRF}\label{sec:fire-budget}\label{sec:fire}\label{si-sec:fire}
Each interpretable model $m\in\{\mathrm{CART},\mathrm{SRF}\}$ in the Arbitrator has a set of rules $\mathcal{R}_m$ over the donor feature space $\mathcal{X}$. Each rule $r\in\mathcal R_m$ is a conjunction of split conditions that defines a region of $\mathcal X$, and we write $x\in r$ when a kidney with features $x$ satisfies it (e.g., donor age $>50$ years \textit{and} KDPI $\le60\%$ \textit{and} terminal creatinine $>1.5$~mg/dL \textit{and} a history of hypertension). For each rule, we attach a 99\% confidence interval $[L_r,U_r]$ for the true early-acceptance (EM) rate $p_r$ of kidneys satisfying $r$, computed from validation-set kidneys. For a kidney $x$, model $m$ combines the intervals of the rules it matches, $A_m(x)=\{r\in\mathcal R_m:x\in r\}$, into a single interval $[L_m(x),U_m(x)]$. This section describes how those intervals are constructed; \S~\ref{sec:arb} describes how they are used to make routing decisions.

The two models share three design choices. First, rules are learned from training outcomes only. Second, each rule's EM rate and its 99\% Wilson interval are estimated from the independent validation set, so the same outcomes are never used both to form a rule and to measure its uncertainty. Third, a kidney not covered by any rule with validation support receives the uninformative interval $[0,1]$. The models differ in how rules are formed. CART partitions the feature space into disjoint terminal leaves, so every kidney falls in exactly one leaf (\S~\ref{sec:cart-uncert}). A Random Forest has far too many leaves across its trees to audit individually. We therefore use Fast Interpretable Rule Extraction (FIRE)~\cite{liu2023fire} to select a sparse subset of the forest's leaves as rules, and we denote the resulting model SparseRF (\S~\ref{sec:fire-uncert}). Unlike CART leaves, SparseRF rules can overlap, so a kidney may match several rules; we combine their intervals into a single conservative interval, the consensus envelope. Furthermore, due to sparsification, it is also possible for an organ not to belong in \textit{any} of the sparsified rules generated by FIRE.

\subsection{CART Wilson intervals}\label{sec:cart-uncert}
For a rule $r$, let $n_r$ and $k_r$ be validation kidney support and early-acceptance count. With $\widehat p_r=k_r/n_r$, $z=z_{1-\alpha/2}$ and $\alpha=0.01$, the interval is~\cite{wilson1927probable}
\begin{equation}\label{eq:si-wilson}
[L_r,U_r]=\frac{\widehat p_r+z^2/(2n_r)\ \pm\ z\sqrt{\widehat p_r(1-\widehat p_r)/n_r+z^2/(4n_r^2)}}{1+z^2/n_r}.
\end{equation}
CART's rules are its terminal leaves, which are disjoint, so $A_{\rm CART}(x)$ contains exactly one rule $r$ and $[L_{\rm CART}(x),U_{\rm CART}(x)]=[L_r,U_r]$. We would assign the full $[0,1]$ to a leaf with no validation support, but for this study, there were no such cases.  Wilson intervals quantify uncertainty in group-level event rates but do not guarantee individual outcomes, and dependence between kidneys from the same donor means that there is can be some mild violation of independence (although this is mitigated for leaves with large enough sample sizes).

\subsection{SparseRF: selecting rules with FIRE}\label{sec:fire-uncert}
Each leaf of the fitted Random Forest is a conjunction of split conditions (e.g., KDPI $\le60\%$ and terminal creatinine $\ge1.635$~mg/dL) and therefore defines a candidate rule, but the forest has far too many leaves to audit. We use Fast Interpretable Rule Extraction (FIRE)~\cite{liu2023fire}, in its sparsity-only form, to extract a small subset of these leaves: FIRE fits a sparse weighted combination of leaf indicators to the training EM labels, and its penalty trades sparsity (i.e., maintaining fewer rules), against coverage (i.e., more kidneys are matched by at least one selected rule). We cap the selection at 128 rules, a fixed design choice made to give SparseRF sufficient coverage (\S\ref{sec:si-budget-sensitivity} compares other caps); with this cap, FIRE selects 80 rules, which form SparseRF's rule set $\mathcal R_{\rm SRF}=\widehat{\mathcal S}$. Neither validation nor test outcomes are used in this selection, and the FIRE coefficients determine only which rules are kept: each selected rule's uncertainty comes from its validation Wilson interval (\S\ref{sec:si-envelope}).

\subsection{SparseRF uncertainty: the consensus envelope}\label{sec:si-envelope}
For a kidney with features $x$, let $A_{\rm SRF}(x)=\{r\in\widehat{\mathcal S}:x\in r\}$ be the selected rules it matches (so $\mathcal R_{\rm SRF}=\widehat{\mathcal S}$), and let $A_s(x)=\{r\in A_{\rm SRF}(x):n_r>0\}$ be those with validation support, where $n_r$ is the number of validation kidneys matching $r$ and $k_r$ the number of them that are EM. Each supported rule receives the 99\% Wilson interval $[L_r,U_r]$ of Eq.~\eqref{eq:si-wilson}, computed from $(n_r,k_r)$. SparseRF's interval for $x$ is the envelope of these intervals:
\begin{equation}\label{eq:si-envelope}
[L_{\rm SRF}(x),U_{\rm SRF}(x)] =
\begin{cases}
[0,1],&A_s(x)=\varnothing,\\
[\min_{r\in A_s(x)}L_r,\max_{r\in A_s(x)}U_r],&A_s(x)\ne\varnothing.
\end{cases}
\end{equation}
A matching rule with no validation support therefore does not contribute, and a kidney matched only by such rules receives $[0,1]$.
SparseRF therefore certifies an action for $x$ exactly when every supported rule that $x$ matches certifies that same action (\S\ref{sec:arb}); otherwise, including when $A_s(x)=\varnothing$, it abstains. Each rule's interval has nominal 99\% coverage on its own, but the envelope carries no simultaneous-coverage guarantee across overlapping rules, and adding rules does not necessarily increase certification rates, because a new overlapping rule can contribute an interval on the other side of $\tau_\beta$.

\subsection{SparseRF rule-budget trade-offs}\label{sec:si-budget-sensitivity}
We compare prespecified budgets $K\in\{8, 16, 32, 64, 128, 256\}$ on the same pooled donor split, fitted forest, and $\beta$-grid. These budgets represent the \textit{maximum} number of selected rules under the sparsification; they are not a required number of rules. Our paper uses a maximum budget of 128 to have a richer bank of rules that support higher levels of SparseRF certification.

\begin{table}[!htb]\centering\footnotesize
\caption{Matched SparseRF rule budgets at $\beta=0.5$. SparseRF and Arbitrator (Arb.) percentages measure certification and rule resolution respectively. Gap is average cost minus the fixed XGBoost benchmark. $K=256$ selects the same 80 rules as $K=128$ and is omitted.}\label{tab:si-fire-budget}
\setlength{\tabcolsep}{4pt}
\begin{adjustbox}{max width=\linewidth}
\begin{tabular}{rrrrrrrr}\toprule
$K$ & Rules & Trees & \shortstack{SparseRF\\cert. (\%)} & \shortstack{Arb.\\(\%)} & Deferred & \shortstack{Average\\cost} & \shortstack{Cost gap\\to XGB}\\\midrule
8 & 7 & 5 & 58.58 & 89.20 & 2,329 & 0.141026 & +0.010290\\
16 & 14 & 7 & 65.96 & 89.86 & 2,187 & 0.141606 & +0.010870\\
32 & 14 & 7 & 65.96 & 89.86 & 2,187 & 0.141606 & +0.010870\\
64 & 34 & 8 & 77.07 & 91.89 & 1,750 & 0.142834 & +0.012098\\
128 & 80 & 9 & 87.34 & 93.54 & 1,393 & 0.142579 & +0.011843\\\bottomrule
\end{tabular}
\end{adjustbox}\end{table}

\begin{figure}[htbp]\centering
\includegraphics[width=.95\linewidth]{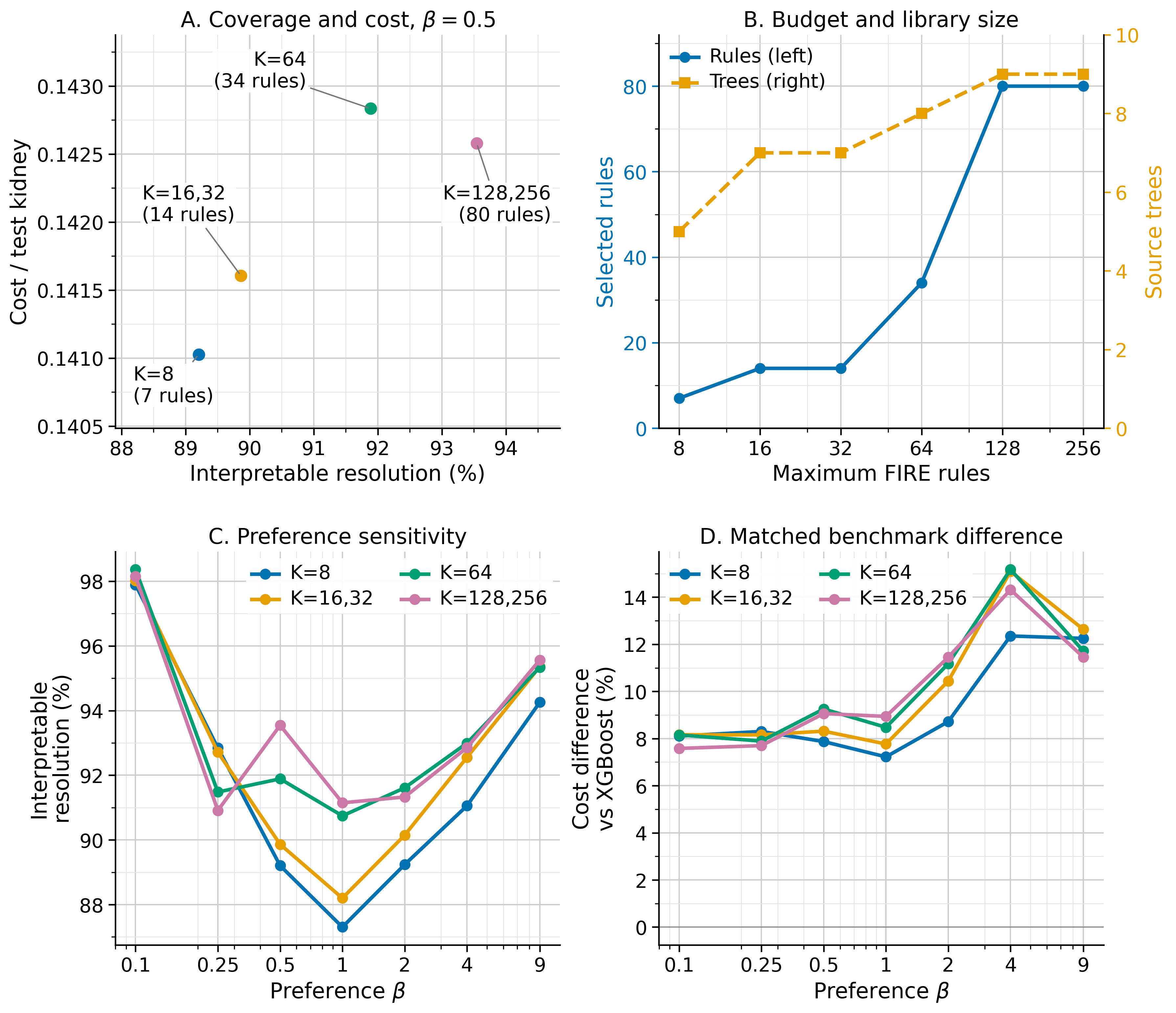}
\caption{SparseRF rule-budget comparison for $K\in\{8, 16, 32, 64, 128, 256\}$ on the same fitted forest and cohort. (A) Average cost and interpretable resolution at $\beta=0.5$. (B) Selected rules and source trees by budget. (C) Interpretable resolution and (D) cost difference relative to XGBoost across $\beta$.}\label{fig:si-fire-tradeoffs}
\end{figure}

Table \ref{tab:si-fire-budget} and Figure~\ref{fig:si-fire-tradeoffs} show that raising the maximum rule budget trades sparsity for coverage. Larger budgets let the sparsification keep more rules from more trees. That raises SparseRF certification and Arbitrator resolution (from 89.2\% to 93.5\% at $\beta=0.5$; panel A), and it substantially reduces the number of deferred kidneys. The cost of that extra coverage is small. Average cost and the gap to XGBoost change only slightly across budgets, and they do not move monotonically in $K$; across the full $\beta$-grid, the cost gap depends far more on $\beta$ than on $K$ (panel D). The budget also acts as a cap rather than a target. SparseRF-16 and SparseRF-32 select the same 14-rule solution, so a larger budget does not always produce a larger rule set, and the rule set saturates at 80 rules from nine trees (panel B). Across preferences, resolution is lowest near $\beta=1$, and larger budgets raise it most for $\beta$ between 0.5 and 2 (panel C). We therefore treat $K$ as a choice between a more compact rule set and broader certification, not as a parameter with a single best value.

\section{Certification and arbitration}\label{sec:arb}\label{sec:si-pipeline-decisions}

\subsection{Decision threshold}
Let $a=+1$ keep a kidney on-sequence and $a=-1$ route it off-sequence, the preferred actions when an organ is EM ($Y=+1$) or non-EM ($Y=-1$), respectively. Let $p(x)=P(Y=+1\mid x)$ be the probability that an organ with features $x$ is EM, and $\widehat p_m(x)$ model $m$'s estimate of it. We treat EM as the positive class: a false positive (FP) keeps a non-EM kidney on-sequence ($a=+1$, $Y=-1$), and a false negative (FN) routes an EM kidney off-sequence ($a=-1$, $Y=+1$). Suppose the cost for a false positive (i.e., keeping a non-EM kidney on-sequence) is 1 and the cost for a false negative (i.e., moving an EM kidney off-sequence) is $\beta \geq 0$. Then, the expected costs for keeping the kidney on-sequence or routing it off-sequence, respectively, are
\begin{equation}
C_{+1}(p)=1-p,\qquad C_{-1}(p)=\beta p.
\end{equation}
The optimal threshold $\tau_\beta$ to minimize the decision costs from expected false positives and negatives is given by
\begin{equation}
\tau_\beta=\frac{1}{1+\beta},\qquad a^*(p)=2 \cdot \mathbf{1}\{p\ge\tau_\beta\}-1.
\end{equation}
We evaluate a policy by its average decision cost, $(\mathrm{FP}+\beta\,\mathrm{FN})/N$, where FP and FN are the numbers of false positives and false negatives out of the full set of $N$ test organs. We report the prespecified grid $\beta\in\{0.1, 0.25, 0.5, 1, 2, 4, 9\}$; results below use $\beta=0.5$, for which $\tau_{0.5}=2/3$.

\subsection{Decision certification under uncertainty}
Each interpretable model $m$ used by the Arbitrator supplies a validation-supported Wilson interval $[L_m(x),U_m(x)]$, constructed as described in \S~\ref{sec:fire-budget}. For a model $m$ making a prediction on organ with feature vector $x$, we certify an on-sequence decision ($a_m(x)=+1$) if $L_m(x)>\tau_\beta$, off-sequence ($a_m(x)=-1$) if $U_m(x)<\tau_\beta$, and abstain ($a_m(x)=0$) otherwise. Certification is therefore a \textit{model-level assessment} of whether the same action is optimal throughout the entire uncertainty interval. In our two-model Arbitrator framework there are four possible \textit{certification pathways} across CART and SparseRF: \emph{dual} (both models certify the same action), \emph{CART-only} (CART certifies and SparseRF abstains), \emph{SparseRF-only} (SparseRF certifies and CART abstains), or \emph{deferred} (both abstain, or they certify opposite actions). The Arbitrator reconciles the possibly different model certifications to decide whether the organ should be routing off-sequence, kept on-sequence, or deferred for expert feedback in a process called \textit{resolution}.

The main text uses the \emph{single certification} resolution policy, which resolves dually certified kidneys and both single-model pathways. Here we also consider a stricter \emph{dual certification} policy, which resolves only dually certified kidneys, to show what single-model resolution adds in terms of coverage.
Under either policy, all non-resolved kidneys are deferred to organ-level review. Because that review is not observed, our retrospective evaluation assigns each deferred kidney the calibrated XGBoost decision at $\tau_\beta$ as a stand-in for an expert reviewer, so every kidney receives a decision, but only resolved kidneys receive an interpretable one. Deferral at this stage reflects statistical uncertainty or disagreement; a separate, later rule-level deferral can arise when an audited rule is not recommended for automation based on clinical research or policy (\S~\ref{sec:si-audit}).

Table~\ref{tab:si-routing-status} assigns each test kidney at $\beta=0.5$ to one of the four resolution pathways. The table compares resolution and average cost under the two policies. The 3,060 total organs that are certified only by CART or SparseRF account for the gap between the two policies' resolution totals (20,181 versus 17,121), and their downstream average cost.
\begin{table}[!htb]\centering\small
\caption{Resolution pathways at $\beta=0.5$ for the 21,574 test kidneys. Dual resolution requires both models to certify the same action; single-model resolution requires one model to certify while the other abstains.}\label{tab:si-routing-status}\label{tab:si-heatmap}\label{tab:si-deferral}
\setlength{\tabcolsep}{5pt}
\renewcommand{\arraystretch}{1.15}
\begin{tabular*}{\linewidth}{@{\extracolsep{\fill}}lrrrrr@{}}\toprule
 & & \multicolumn{2}{c}{Resolution policy} & \multicolumn{2}{c}{Average cost}\\
\cmidrule(lr){3-4}\cmidrule(l){5-6}
Pathway & Kidneys (\%) & Single & Dual & Single & Dual\\\midrule
Dual & 17,121 (79.36) & Resolved & Resolved & 0.1136 & 0.1136\\
CART-only & 1,689 (7.83) & Resolved & Deferred & 0.2866 & 0.2238\\
SparseRF-only & 1,371 (6.35) & Resolved & Deferred & 0.2666 & 0.2221\\
Deferred & 1,393 (6.46) & Deferred & Deferred & 0.2021 & 0.2021\\\midrule
Total & 21,574 (100.00) &  &  & 0.1426 & 0.1348\\\bottomrule
\end{tabular*}
\par\smallskip
\end{table}
The two resolution policies trade coverage against corroboration and average decision cost. Under dual certification, the Arbitrator acts only when both models independently certify the same action, which holds for 17,121 kidneys (79.36\%) at an average cost of 0.1136. Corroboration is rarely the bottleneck: when both models certify, they agree on 98.0\% of kidneys, so most kidneys outside the dual pathway are there because one or both models abstain, not because the models disagree. Single certification resolves another 3,060 kidneys (14.18\%), raising rule-based resolution to 93.54\% and leaving 6.46\% for organ-level review instead of 20.64\%. The two models are complementary on these kidneys: most CART-only resolutions (67.08\%) involve kidneys that no supported SparseRF rule covers, while every SparseRF-only resolution involves a kidney in a CART leaf whose interval is inconclusive. Because the two policies make identical decisions on every other resolved kidney, these 3,060 resolutions account for the entire difference in overall cost. Relative to XGBoost applied to every kidney (0.1307), organs receiving dual certification cost 3.1\% more and organs receiving either dual or single certification cost 9.1\% more. Each single-model decision rests on one model's evidence while the other model's evidence is absent or inconclusive. Whether accepting these single-certified decisions is worthwhile therefore depends on the value placed on interpretable coverage and on how their rules fare under audit.

Table~\ref{tab:si-rule-atlas} shows example kidney profiles from each resolution pathway at $\beta=0.5$. In the dual-certified profiles (A, B), both intervals lie on the same side of $\tau_\beta=2/3$. In the single-certified profiles, one model certifies while the other abstains: no supported SparseRF rule covers profiles C and D, whereas the CART interval crosses the threshold in E and F. Neither model certifies profile G, so the Arbitrator defers it for organ-level review, represented here by XGBoost. Profile D shows CART region C35 (Table~\ref{tab:si-audit-examples}) for kidneys that no supported SparseRF rule covers. It is difficult to explicitly characterize organs for which CART is better suited versus SparseRF; however, we observe that single-certified and deferred organs have more of a ``mix'' of positive and negative feature coonjunctions in their rules than the dual-certified rules, which are more intuitive.

\begin{table}[!htb]\centering
\caption{Example certification profiles at $\beta=0.5$. OS denotes on-sequence and OOS off-sequence placement. A model certifies OS when its entire 99\% confidence interval (CI) is above $\tau_\beta=2/3$, certifies OOS when it is below, and abstains otherwise. The number of test kidneys matching each profile is given by $n$. C identifies the CART leaf and F a selected SparseRF rule. Donor features are abbreviated observed-value summaries; missing-value alternatives are omitted. \textit{Cr, terminal creatinine; GS, glomerulosclerosis; IF, interstitial fibrosis; BUN, blood urea nitrogen; HCV NAT, hepatitis C nucleic acid test.}}\label{tab:si-rule-atlas}\label{fig:si-rule-atlas}
\begingroup
\footnotesize
\setlength{\tabcolsep}{4pt}
\renewcommand{\arraystretch}{1.25}
\renewcommand{\tabularxcolumn}[1]{m{#1}}
\newcommand{\atlasCell}[1]{\parbox[c]{\linewidth}{\vspace*{3pt}\raggedright #1\par\vspace*{3pt}}}
\begin{tabularx}{\linewidth}{@{}>{\raggedright\arraybackslash}m{2.7cm} >{\raggedright\arraybackslash}X >{\raggedright\arraybackslash}m{1.9cm} >{\centering\arraybackslash}m{.9cm} >{\centering\arraybackslash}m{2.0cm} >{\centering\arraybackslash}m{2.0cm}@{}}
\toprule
\shortstack[l]{\textbf{Certification}\\\textbf{pathway/action}} & \textbf{Donor features} & \textbf{Rule(s)} & $n$ & \shortstack{\textbf{CART}\\\textbf{99\% CI}} & \shortstack{\textbf{SparseRF}\\\textbf{99\% CI}}\\
\midrule
\atlasCell{A. Dual-certified\\\textbf{OOS}} & \atlasCell{KDPI $>90.5\%$; GS $>20\%$; vascular changes $\ge10\%$.} & C164,\allowbreak{} F153\_400 & 158 & \shortstack{0.032--0.067\\OOS} & \shortstack{0.023--0.057\\OOS}\\
\midrule[0.3pt]
\atlasCell{B. Dual-certified\\\textbf{OS}} & \atlasCell{KDPI $\le41.5\%$; Cr $\le2.595$ mg/dL; IF $<10\%$; GS $\le10\%$; lipase $>1290.5$ U/L; no kidney biopsy; HCV NAT negative.} & C19,\allowbreak{} F25\_13,\allowbreak{} F81\_9,\allowbreak{} F164\_278,\allowbreak{} F206\_243 & 2,077 & \shortstack{0.918--0.939\\OS} & \shortstack{0.888--0.932\\OS}\\
\midrule
\atlasCell{C. CART-only\\\textbf{OOS}} & \atlasCell{$78.5\%<$ KDPI $\le90.5\%$; GS $\le10\%$; recovery year through 2019; Cr $\le2.295$ mg/dL; continued smoking not recorded; BUN $>16.5$ mg/dL.} & C130 & 86 & \shortstack{0.499--0.663\\OOS} & \shortstack{0.000--1.000\\defer}\\
\midrule[0.3pt]
\atlasCell{D. CART-only\\\textbf{OS}} & \atlasCell{$41.5\%<$ KDPI $\le60.5\%$; Cr $\le1.675$ mg/dL; IF $<10\%$; GS $\le10\%$; no coronary angiogram; recovery year after 2019.} & C35 & 131 & \shortstack{0.682--0.764\\OS} & \shortstack{0.000--1.000\\defer}\\
\midrule
\atlasCell{E. SparseRF-only\\\textbf{OOS}} & \atlasCell{$60.5\%<$ KDPI $\le78.5\%$; GS $>10\%$; kidney biopsy not recorded as No; IF $<10\%$; Cr $\le1.895$ mg/dL; age $\le61.5$ years; vascular changes $<10\%$.} & C96,\allowbreak{} F153\_374 & 45 & \shortstack{0.505--0.675\\defer} & \shortstack{0.311--0.414\\OOS}\\
\midrule[0.3pt]
\atlasCell{F. SparseRF-only\\\textbf{OS}} & \atlasCell{$27.5\%<$ KDPI $\le43.5\%$; $2.595<$ Cr $\le3.765$ mg/dL; sodium $>142.5$ mEq/L.} & C49,\allowbreak{} F25\_59 & 72 & \shortstack{0.557--0.786\\defer} & \shortstack{0.788--0.882\\OS}\\
\midrule
\atlasCell{G. Deferred} & \atlasCell{KDPI $\le41.5\%$; Cr $\le2.595$ mg/dL; IF $\ge10\%$.} & C21 & 112 & \shortstack{0.486--0.710\\defer} & \shortstack{0.000--1.000\\defer}\\

\bottomrule
\end{tabularx}
\endgroup
\end{table}



\subsection{Potentially saved organs}\label{sec:si-potentially-saved}
Table~\ref{tab:si-potentially-saved}
reports how many kidneys in the two low-KDPI populations defined in \S~\ref{sec:si-cutoff} are identified for OOS placement at $\beta=0.5$ by the single certification Arbitrator and by XGBoost applied to every kidney.
We use this as a proxy for the percent of transplantable organs that could be saved by our methodology. The Arbitrator's identifications trace to 113 certified rules that can be audited before automation (\S~\ref{sec:si-audit}).

\begin{table}[htb]\centering\small
\caption{Trade-off between interpretable resolution and potentially saved organs at $\beta=0.5$. Arbitrator denotes the single certification policy used in the main paper; the dual-certification-only policy resolves a kidney only when both models certify the same action. Interpretable percentages use all 21,574 test kidneys, and organs resolved counts decisions supported by interpretable rules. Both Arbitrator policies use XGBoost for deferred kidneys. There are 4,944 low-KDPI non-EM and 2,738 low-KDPI historically discarded kidneys in the test data.}\label{tab:si-potentially-saved}
\setlength{\tabcolsep}{7pt}
\renewcommand{\arraystretch}{1.2}
\begin{tabularx}{\linewidth}{@{}Xrrrr@{}}\toprule
Policy
& \shortstack{Interpretable\\(\%)}
& \shortstack{Organs\\resolved}
& \shortstack{Low-KDPI,\\non-EM\\identified}
& \shortstack{Low-KDPI,\\discarded\\identified} \\\midrule
XGBoost-only & 0.0 & --- & 3,300 (66.7\%) & 2,079 (75.9\%) \\
Arbitrator & 93.5 & 20,181 & 3,068 (62.1\%) & 1,920 (70.1\%) \\
Arbitrator (dual-certification only) & 79.4 & 17,121 & 3,188 (64.5\%) & 1,992 (72.8\%) \\\bottomrule
\end{tabularx}
\end{table}

\section{Rule audit: decide, defer, and candidate deviation}\label{sec:si-audit}
This section describes the exploratory LLM-assisted audit of the Arbitrator's certified rules: how we performed it (\S\ref{sec:si-audit-methods}), what it recommended (\S\ref{sec:si-audit-results}), how a hypothetical policy would apply those recommendations to kidneys (\S\ref{sec:si-audit-policy}), and selected rules in detail (\S\ref{sec:si-audit-sample}).

\subsection{Audit procedure}\label{sec:si-audit-methods}
We audited the 113 resolved contributing rules at $\beta=0.5$: 35 condensed\footnote{We condense a CART subtree to the parent node for all leaves which stare the same certification status.  For abstaining leaves, the tentative point-prediction action must also agree to be condensed. The Wilson interval is the envelope of the original leaf intervals, and validation supports are summed} CART regions and 78 selected SparseRF rules. The 19 CART and two SparseRF regions deferred during the resolution stage are not audited.
We asked an OpenAI GPT-6-based Codex agent to identify 10--20 papers to inform assessments of certified Arbitrator rules. The instructions defined ``decide'' as retaining an action aligned with clinical literature, ``defer'' as requesting physician review when evidence was mixed, and ``deviate'' as proposing the opposite action in light of new research or policy. Our instructions encouraged permissive decide and deviate judgments. We did not pre-specify literature topics, search terms, study eligibility criteria, or a means of determining how to rate the evidence. The agent selected topics from the rule features and supplied context, assembling 18 research papers and two clinical guidance papers comprising topics like AKI and recovery, biopsy findings, KDPI, expedited allocation, equity, HCV treatment, and machine perfusion, listed below. This was an exploratory search. Prompt refinement and clinical validation remains future work.
\begin{multicols}{2}
\begin{enumerate}
\item Liu et al. (2020)~\cite{Liu2020}.
\item Wen et al. (2024)~\cite{Wen2024}.
\item Na et al. (2026)~\cite{Na2026}.
\item Husain et al. (2025)~\cite{Husain2025}.
\item Rao et al. (2009)~\cite{rao2009}.
\item Massie et al. (2014)~\cite{massie2014}.
\item Lentine et al. (2019)~\cite{audit_Lentine2019}.
\item Cheungpasitporn et al. (2020)~\cite{Cheungpasitporn2020}.
\item Jadlowiec et al. (2023)~\cite{audit_Jadlowiec2023}.
\item Malinoski et al. (2023)~\cite{Malinoski2023}.
\item Hall et al. (2019)~\cite{audit_Hall2019}.
\item Liyanage et al. (2025)~\cite{audit_Liyanage2025}.
\item King et al. (2022)~\cite{king2022}.
\item Larpparisuth et al. (2023)~\cite{Larpparisuth2023}.
\item Sise et al. (2022)~\cite{audit_Sise2022}.
\item King et al. (2021)~\cite{audit_King2021}.
\item Yu et al. (2026)~\cite{Yu2026}.
\item Paulus et al. (2026)~\cite{audit_Paulus2026}.
\item OPTN: KDPI refit~\cite{audit_OPTN2025}.
\item AASLD/IDSA: HCV guidance \cite{AASLDIDSAHCV}.
\end{enumerate}
\end{multicols}

After inspecting the complete rule profiles, validation intervals, and selected literature, the agent wrote an assessment-assembly script. It supplied case-specific rationales for the eight highlighted deviation candidates and several defer profiles, then used profile-group assignments for the remaining rules. In general we observed the following trends from the auditor. On-sequence profiles generally retained their action, OOS profiles with a KDPI lower bound of at least 78\% or a condition requiring fibrosis or vascular changes above a specified severity threshold generally retained theirs, and other OOS profiles generally deferred.

\subsection{Audit recommendations}\label{sec:si-audit-results}
Table~\ref{tab:si-audit} summarizes the provisional recommendations: the audit retained 62 rule actions (decide), referred 43 rules for physician review (defer), and proposed reconsidering eight (deviate). All eight deviation candidates propose considering on-sequence placement instead of the learned off-sequence action. Six concern creatinine/possible acute kidney injury (AKI), and two concern a sclerosis threshold.

\begin{table}[!htb]\centering\small
\caption{Provisional recommendations from an LLM audit of the 21,574-kidney test cohort. Organ counts are unique kidneys matching at least one audited rule in that category.}\label{tab:si-audit}
\setlength{\tabcolsep}{4pt}
\begin{adjustbox}{max width=\linewidth}
\begin{tabular}{lrrrrrr}\toprule
& \multicolumn{2}{c}{CART} & \multicolumn{2}{c}{SparseRF} & \multicolumn{2}{c}{Combined}\\
\cmidrule(lr){2-3}\cmidrule(lr){4-5}\cmidrule(lr){6-7}
Recommendation & Rules & Organs & Rules & Organs & Rules & Organs\\\midrule
Decide: provisionally retain action & 18 & 15,903 & 44 & 16,219 & 62 & 17,459\\
Defer: physician review needed & 11 & 2,451 & 32 & 5,130 & 43 & 5,802\\
Deviate: reconsider opposite action & 6 & 806 & 2 & 217 & 8 & 921\\
\midrule
All audited rules (union) & 35 & 19,160 & 78 & 19,504 & 113 & 20,726\\
\bottomrule
\end{tabular}
\end{adjustbox}
\end{table}

\subsection{Hypothetical audit policy}\label{sec:si-audit-policy}
To estimate how many kidneys could still be potentially automated after the audit, we apply a hypothetical policy. A decide recommendation retains its rule's learned action, a deviate recommendation supplies the opposite action, and a defer recommendation supplies no action. Only rules from a model that certified the kidney's action count toward the automation decision. SparseRF contributes only if every matching contributing rule supplies the same certified action. If any contributing SparseRF rule offers the ``defer'' recommendation or conflicts with another rule, SparseRF cannot recommend an audited decision unless there is an supporting CART rule for a kidney that was originally dually certified. A resolved kidney remains eligible for automation if at least one model rule is clean under the audit; otherwise it is deferred. Under this policy, 17,471 of the 20,181 originally resolved kidneys (86.6\%) remain eligible, comprising 13,508 on-sequence and 3,963 off-sequence actions. Of the 2,710 that defer, 13 have conflicting component actions and 2,697 have no usable action.

\begin{table}[!htb]\centering\small
\caption{Organ-level audit outcomes by original certification pathway. Retained organs remain eligible for automation under the hypothetical audit policy. Deferral percentages use the organs audited in each row. Overall combines Dual and Single; the two indented rows partition Single and must not be added to it. The 1,393 organs deferred before audit are excluded.}\label{tab:si-audit-policy-pathways}
\setlength{\tabcolsep}{8pt}
\renewcommand{\arraystretch}{1.15}
\begin{tabularx}{\linewidth}{Xrrr}\toprule
Original certification & \shortstack{Organs\\audited} & \shortstack{Retained for\\automation} & \shortstack{Deferred\\$n$ (\%)}\\\midrule
Dual & 17,121 & 15,401 & 1,720 (10.05\%)\\
Single & 3,060 & 2,070 & 990 (32.35\%)\\
\hspace*{1em}CART only & 1,689 & 1,230 & 459 (27.18\%)\\
\hspace*{1em}SparseRF only & 1,371 & 840 & 531 (38.73\%)\\
\midrule
Overall & 20,181 & 17,471 & 2,710 (13.43\%)\\
\bottomrule
\end{tabularx}
\end{table}

Table~\ref{tab:si-audit-policy-pathways} breaks final deferral down by original resolution pathway. Single-model resolutions defer at 32.35\% (990 of 3,060), compared with 10.05\% (1,720 of 17,121) for dual resolutions. The CART-only and SparseRF-only rates are 27.18\% and 38.73\%, respectively. At the rule level, 35 of the 83 distinct rules contributing to single-certified kidneys received a defer recommendation, compared with 43 of the 113 contributing to dually certified kidneys, noting that a rule can contribute to both pathways.

\subsection{Supported deviation candidates and comparison profiles}\label{sec:si-audit-sample}
Table~\ref{tab:si-audit-examples} presents all eight deviation candidates and three selected decide/defer comparisons. CART identifiers denote condensed regions, and SparseRF identifiers denote source tree and node. We give a brief analysis of these selected rules from the audit.

\begin{table}[!htbp]\centering\footnotesize
\caption{Audited candidate and comparison rules. CI is the Wilson interval for historical early acceptance computed from the validation data. Descriptions summarize observed-value conditions joined by ``AND.'' OS denotes on-sequence, OOS denote off-sequence, and arrows show proposed deviations. Test supports can overlap. \textit{Abbreviations: KDPI: historical Kidney Donor Profile Index; Cr: terminal creatinine (mg/dL); IF: interstitial fibrosis; GS: glomerulosclerosis; BUN: blood urea nitrogen (mg/dL); Na: serum sodium (mEq/L); age in years; CA: California.}}\label{tab:si-audit-examples}
\setlength{\tabcolsep}{3pt}
\renewcommand{\arraystretch}{1.12}
\begin{tabularx}{\linewidth}{@{}l>{\raggedright\arraybackslash}Xcrr>{\raggedright\arraybackslash}p{.18\linewidth}@{}}\toprule
Rule & Condensed description & CI & \shortstack{Val.\\$n$} & \shortstack{Test\\$n$} & Auditor recommendation\\\midrule
C36 & $41.5\%<\mathrm{KDPI}\le60.5\%$; $1.675<\mathrm{Cr}\le2.595$; IF $<10\%$; GS $\le10\%$; no coronary angiogram; recovery after 2019
& $[.377,.645]$ & 86 & 97 & Deviate: \textbf{OOS}$\rightarrow$OS \cite{Liu2020,Wen2024,Na2026}\\
C51 & KDPI $\le43.5\%$; Cr $>3.765$; cause of death not recorded as anoxia
& $[.158,.534]$ & 176 & 222 & Deviate: \textbf{OOS}$\rightarrow$OS \cite{Liu2020,Wen2024,Na2026}\\
F25\_70 & Age $\le50.5$; KDPI $\le43.5\%$; GS $\le10\%$; Cr $>3.765$; cause of death not recorded as anoxia; state not recorded as CA; glomerular count $\le27.5$
& $[.071,.317]$ & 57 & 51 & Deviate: \textbf{OOS}$\rightarrow$OS \cite{Liu2020,Wen2024,Na2026}\\
C58 & KDPI $\le43.5\%$; Cr $>3.765$; anoxia; age $>39.5$
& $[.399,.666]$ & 86 & 102 & Deviate: \textbf{OOS}$\rightarrow$OS \cite{Liu2020,Wen2024,Na2026}\\
C62 & $43.5\%<\mathrm{KDPI}\le60.5\%$; Cr $>5.025$; GS $\le15\%$
& $[.391,.650]$ & 92 & 44 & Deviate: \textbf{OOS}$\rightarrow$OS \cite{Liu2020,Wen2024,Na2026}\\
F164\_72 & Biopsy type recorded; KDPI $\le67.5\%$; BUN $>34.5$; IF $<10\%$; GS $\le15\%$; Na $>142.5$; Cr $>3.205$; age $>42.5$
& $[.306,.513]$ & 143 & 167 & Deviate: \textbf{OOS}$\rightarrow$OS \cite{Liu2020,Wen2024,Na2026}\\
C20 & KDPI $\le41.5\%$; Cr $\le2.595$; IF $<10\%$; GS $>10\%$
& $[.410,.653]$ & 105 & 118 & Deviate: \textbf{OOS}$\rightarrow$OS \cite{Husain2025,Cheungpasitporn2020}\\
C37 & $41.5\%<\mathrm{KDPI}\le60.5\%$; Cr $\le2.595$; IF $<10\%$; GS $>10\%$
& $[.156,.656]$ & 241 & 223 & Deviate: \textbf{OOS}$\rightarrow$OS \cite{Husain2025,Cheungpasitporn2020}\\\midrule
C35 & $41.5\%<\mathrm{KDPI}\le60.5\%$; Cr $\le1.675$; IF $<10\%$; GS $\le10\%$; no coronary angiogram; recovery after 2019
& $[.682,.764]$ & 788 & 756 & Decide: OS \cite{rao2009,Liu2020}\\
F25\_109 & Age $\le50.5$; GS not recorded; Cr $>2.635$; KDPI $>49.5\%$; state not recorded as CA
& $[.149,.298]$ & 196 & 213 & Defer: N/A \cite{rao2009,Liu2020}\\
C133 & KDPI $>90.5\%$; GS $\le10\%$
& $[.052,.600]$ & 470 & 459 & Decide: \textbf{OOS} \cite{rao2009}\\\bottomrule
\end{tabularx}
\par\smallskip\raggedright
\end{table}

\textbf{F25\_70}. This rule proposes OS despite a validation early-acceptance interval of $[0.071,0.317]$ from 57 kidneys. Excluding the high creatine, all other metrics look favorable: relatively low age, low KDPI, and moderately low levels of sclerosis. These latter factors motivate consideration of reversible AKI, which, if true, should not lead to significantly lower graft success rates~\cite{Liu2020,Wen2024,Na2026}. Thus, this kidney may be a candidate for early-acceptance if (a) the research is validated and (b) clinician acceptance behaviors align with this new research.

\textbf{C36 versus C35.} These rules are nearly identical with the exception of creatinine values, and this difference drives their certified decisions as being OOS and OS, respectively. These profiles otherwise share a moderate KDPI range, limited recorded fibrosis/sclerosis, no coronary angiogram and recovery after 2019, but split at creatinine 1.675 mg/dL. Their validation intervals support opposite learned actions, but AKI evidence motivates reviewing whether an isolated creatinine increase warrants expedited placement in the presence of new research~\cite{Liu2020,Wen2024,Na2026}.

\textbf{C20 and C37.}
Both propose OS reconsideration for a learned GS $>10\%$ branch. The GS scale admits categorical values of 0--5\%, 6--10\%, 11--15\%, 16--20\%, and $>20\%$, so the sclerosis threshold appears to be too general to make the immediate OOS action. Prior studies restricted to KDPI above 85\% cannot validate this threshold in lower-KDPI kidneys~\cite{Husain2025,Cheungpasitporn2020}.

\textbf{F25\_109 and C133.} These rules show why defer and decide differ. F25\_109 has a very low upper bound on its confidence interval at 0.298, but the absence of GS findings and an upper bound on KDPI renders the rule too general to immediately recommend OOS. In contrast, C133 has a higher upper bound of 0.600, but has observed KDPI above 90.5\%, which is likely sufficiently high to warrant OOS, even if the level of sclerosis is moderately low~\cite{rao2009,Liu2020}.

\end{document}